\documentclass[PHME, 2024]{PHMSociety}

\usepackage{graphicx}
\usepackage{amsmath}
\usepackage{lipsum}
\usepackage{algorithm}
\usepackage{mathtools}
\usepackage{multirow}
\usepackage{dsfont}
\usepackage{subfigure}
\usepackage{threeparttable} 
\usepackage{booktabs} 
\usepackage[table]{xcolor}
\usepackage{hhline} 
\usepackage{apacite}
\graphicspath{{Figures/}} 

\begin{document}

\title{Towards a Probabilistic Fusion Approach for \\Robust Battery Prognostics}

\author{%
	Jokin Alcibar\authorNumber{1}, Jose I. Aizpurua\authorNumber{2,3}, and Ekhi Zugasti\authorNumber{4}
}
\address{
	\affiliation{{1,2,4}}{Electronics \& Computer Science Department, Mondragon University, Spain}{ 
		{\email{jalcibar@mondragon.edu}}\\ 
		{\email{ezugasti@mondragon.edu}}
		} 
	\tabularnewline 
	\affiliation{3}{Ikerbasque, Basque Foundation for Science, Bilbao, Spain}{ 
		{\email{jiaizpurua@mondragon.edu}}
		} 
}

\maketitle
\pagestyle{fancy}
\thispagestyle{plain}

\phmLicenseFootnote{Jokin Alcibar}

\begin{abstract}

Batteries are a key enabling technology for the decarbonization of transport and energy sectors. The safe and reliable operation of batteries is crucial for battery-powered systems. In this direction, the development of accurate and robust battery state-of-health prognostics models can unlock the potential of autonomous systems for complex, remote and reliable operations. The combination of Neural Networks, Bayesian modelling concepts and ensemble learning strategies, form a valuable prognostics framework to combine uncertainty in a robust and accurate manner. Accordingly, this paper introduces a Bayesian ensemble learning approach to predict the capacity depletion of lithium-ion batteries. The approach accurately predicts the capacity fade and quantifies the uncertainty associated with battery design and degradation processes. The proposed Bayesian ensemble methodology employs a stacking technique, integrating multiple Bayesian neural networks (BNNs) as base learners, which have been trained on data diversity. The proposed method has been validated using a battery aging dataset collected by the NASA Ames Prognostics Center of Excellence. Obtained results demonstrate the improved accuracy and robustness of the proposed probabilistic fusion approach with respect to (i) a single BNN model and (ii) a classical stacking strategy based on different BNNs.


\end{abstract}

\section{Introduction}
\label{sed:Intro}

Batteries are key components in the transition towards a sustainable carbon-free economy. In this transition, the development of remaining useful life (RUL) prediction of batteries is a crucial activity. The accuracy and reliability of the RUL prediction models is essential to build trust in the predictions \cite{liu2023a}. In this context, robust and reliable battery prognostics models support the development of accurate monitoring strategies and cost-effective solutions.

The estimation of the state-of-health (SOH) of batteries is a key activity for the design of RUL prognostics models. SOH-based prognostics models focus on capturing the run-to-failure ageing dynamics and battery health state estimation \cite{TOUGHZAOUI2022}. It is frequently used to determine age-related degradation that reduces energy capacity and rises safety risks, including overheating and explosions~\cite{wang2022}. Therefore, accurate SOH monitoring and forecasting are key activities to design and operate safe, reliable and effective battery-powered systems~\cite{ZHAO2023}.


SOH estimation is an ongoing area of research~\cite{yang2023}. SOH refers to the ratio of the current maximum capacity relative to its original specified capacity ~\cite{zhao2024}. SOH can be quantified through different factors, including resistance and maximum power. However, discharge capacity is the most common definition~\cite{vanem2021}, and this is adopted in this research.

Recent data-driven approaches have focused on modeling the capacity degradation of lithium-ion (Li-ion) batteries. \cite{LEE2023} used convolutional neural network (CNN) to estimate the future SOH value of Li-ion batteries, transforming the capacity degradation data into two-dimensional images. Estimates of the SOH and RUL are commonly found together in the literature. For example, \cite{TOUGHZAOUI2022} developed a CNN-LSTM architecture, and \cite{wei2023} presented a graph CNN complemented by dual attention mechanisms for the estimation of SOH and RUL of batteries. However, due to the variability inherent in battery manufacturing process, it is essential to quantify this uncertainty to ensure robust and reliable prognostics predictions \cite{abdar2021, Nemani2023}.

There are different sources of uncertainty present in the design, operation and maintenance of batteries \cite{hadigol2015}. \cite{zhang2024} introduced a SOH assessment method that estimates uncertainty through the quantile distribution of deep features, which are inferred from a Residual Neural Network (ResNet) architecture. This approach generates SOH values accompanied by confidence intervals. However, the proposed ResNet architecture lacks probabilistic layers, overlooking the uncertainty inherent in the model parameters. \cite{che2024} developed a prognostic framework to assess battery aging, using a CNN-LSTM Bayesian neural network. However, this approach limits the uncertainty to the final dense layers, which are the only components modeled probabilistically. 

With the aim of capturing uncertainty associated with complex processes, recent studies in the broader machine learning (ML) community have focused on ensembles of probabilistic models.  \cite{fan2017} introduced a Bayesian posterior predictive framework for weighting ensemble climate models. \cite{cobb2019} present a new ML retrieval method based on an ensemble of Bayesian Neural Networks (BNNs). In this scenario, the overall output from the ensemble is treated as a Gaussian mixture model. However, models are equally weighted with no adaptation to the observed data. \cite{Zhang2022} present a Bayesian Mixture Neural Network (BMNN) for Li-ion battery RUL prediction. The BMNN framework incorporates a Bayesian Convolutional Neural Network as feature extractor and a Bayesian Long Short-Term Memory to learn degradation patterns over time. However, the absence of a weighted model combination limits the analysis of individual model contributions. 

Alternatively, \cite{bai2023a} described a Bayesian ensemble learning framework that uses gradient boosting by combining multiple Neural Networks trained by Markov Chain Monte Carlo (MCMC) sampling. Finally, \cite{dai2023} demonstrate the robustness of Bayesian fusion by embedding the Monte Carlo fusion framework within a sequential Monte Carlo algorithm.




In this context, inspired by the use of probabilistic ensemble models to capture model uncertainty, the main contribution of this research is the development of a novel probabilistic model fusion approach for battery SOH predictions. Bayesian convolutional neural networks (BCNNs) are used as base models for SOH prediction, and the fusion approach integrates individual BCNN probabilistic predictions. The fusion strategy balances between precision and reliability of individual predictions, adopting an optimal tradeoff between accuracy and uncertainty of predictions through the proposed stacking approach.


The proposed approach has been compared with (i) individual BCNN models and (ii) fusion strategies focused on stacking of BCNN models using point prediction information. Obtained results confirm that the proposed framework infers accurate, well-calibrated, and reliable probabilistic predictions, which improve predictive performance and contribute to estimate uncertainty in a robust and reliable manner in complex data-driven tasks. The proposed approach has been tested and validated with the publicly available NASA's battery dataset \cite{saha2007}.

The remainder of this article is organized as follows. Section~\ref{sec:Data_Method} outlines our probabilistic fusion approach for robust battery prognostics. Section~\ref{sec:Case_study} describes a case study to demonstrate the application of our methodology. Section~\ref{sec:Results} presents and analyzes the results obtained from the case study. Section~\ref{sec:Discussion} discusses the implications of these findings. The article concludes with Section~\ref{sec:Conclusions}, summarizing our main conclusions and suggesting avenues for future work.

\section{Probabilistic Fusion Approach for Robust Battery Prognostics}
\label{sec:Data_Method}

The proposed probabilistic fusion framework integrates BCNNs with probabilistic ensemble strategies. The main objective of the integration is to generate accurate predictions with robust uncertainty quantification, thanks to the uncertainty quantification of Bayesian modelling \cite{blundell2015} and the robustness and accuracy of ensemble strategies \cite{Zhang2022}. 

\begin{figure}[!ht]
    \centering
    \includegraphics[width=0.9\linewidth]{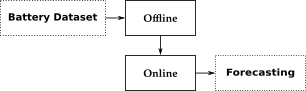}
    \caption{High-level block diagram of the proposed approach.}
    \label{fig:HighLevelConceptual}
\end{figure}

The approach is divided into offline and online stages. Starting from a set of battery datasets, in the offline process, data pre-processing and model training steps are completed. In the online process, trained models are stacked in an ensemble model according to computed weight and stacking criteria. The outcome of the approach is a one-step-ahead probabilistic capacity estimate. Figure~\ref{fig:HighLevelConceptual} shows the high-level block diagram of the proposed approach.

\begin{figure*}
    \centering
    \includegraphics[width=0.99\linewidth]{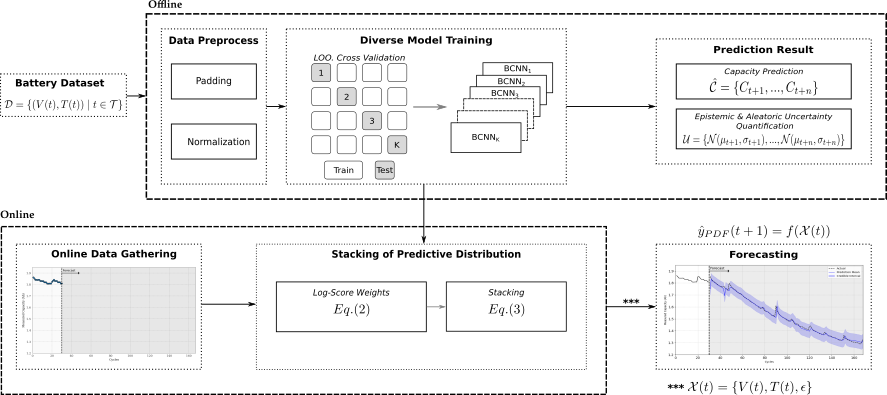}
    \caption{Block diagram of the proposed approach.}
    \label{fig:FlowchartMethodology}
\end{figure*}

The high-level concepts in Figure~\ref{fig:HighLevelConceptual} are implemented through the detailed model architecture shown in Figure~\ref{fig:FlowchartMethodology}.

The base models are BCNN models, which are trained (offline) through a leave-one-out cross validation (LOOCV) process. The probabilistic results of individual BCNN models are aggregated through a stacking process that includes accuracy and uncertainty metrics. In the testing (online) phase, each BCNN model weights are computed using learned models (log-score weights) and the stacking model is designed to combine them and generate a distribution from a mixture model. The following subsections explain in detail the main parts of the approach.

\subsection{Offline Phase}

During the offline phase, starting from a battery dataset with different run-to-failure trajectories on the same type of batteries, different base models are designed through a training strategy which seeks diversity in the training set to develop complementary predictive models.


\subsubsection{Ensemble Base Models: BCNNs}

BCNN models are a Bayesian extension of the classical CNN models to include uncertainty associated with parameter estimation. This requires modification of the classical backpropagation algorithm through Bayesian techniques that involves incorporating uncertainty into the model by treating weights as random variables, and applying variational inference to approximate posterior distributions. This results in a more robust model that predicts the complete probability density function (PDF).

Consequently, BCNN models have been selected to improve the robustness and accuracy of model prediction. To this end, BCNNs make use of probabilistic distributions to model parameters and the uncertainty related to their training process, and prior distributions to incorporate previous knowledge, generate uncertainty estimations and mitigate over-fitting \cite{blundell2015}. In contrast, the classical learning models, e.g. non-Bayesian CNN models, focus on  maximum likelihood estimation (MLE) and they overlook prior and posterior distributions. This leads to increasing error and decreasing model robustness in high uncertainty contexts, e.g. out-of-distribution data or manufacturing drifts.

The proposed approach utilizes data pre-processing techniques to standardize the length of discharge cycles through padding. This technique involves repeating the last discharge value until the desired cycle length is reached, ensuring consistent input dimensions for all models. Additionally, normalization is carried out scaling the discharge values between 0 and 1.

The architecture of the BCNN models is shown in Figure~\ref{fig:BCNN} defined as follows:

\begin{figure*}
    \centering
    \includegraphics[width=0.7\linewidth]{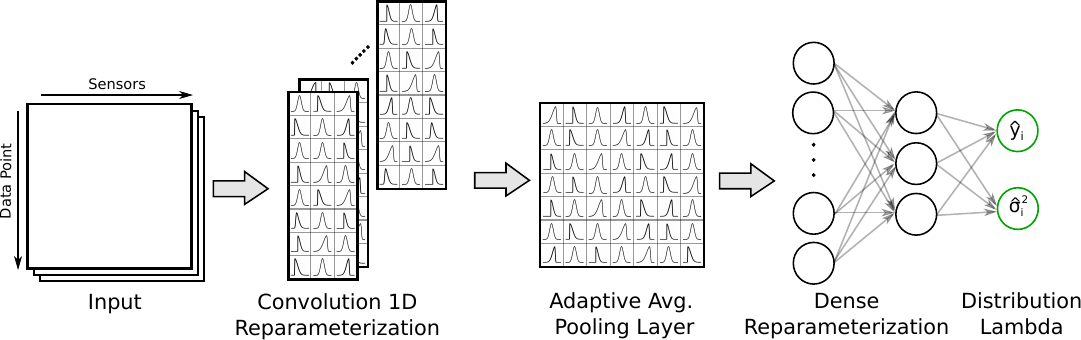}
    \caption{Schematic of the Bayesian convolution neural network.}
    \label{fig:BCNN}
\end{figure*}

\begin{itemize}
    \item Input data: the input data for the BCNN is structured in a tensor format. The rows represent data samples of discharge cycles, and columns that correspond to features, such as the voltage and temperature over time. Notably, the input does not include the current discharge as it remains constant in this scenario.

    \item Convolutional 1D Reparametrization: this layer creates a convolution kernel that is applied to the input data. During the forward pass, kernel and bias parameters are drawn from a Gaussian distribution. It uses the reparameterization estimator to approximate distributions through Monte Carlo trials, integrating over the kernel and bias.

    \item Global Average Pooling 1D: this layer performs average pooling specifically for temporal data. It reduces the spatial dimensions of the input data to a single value per channel by calculating the average over the temporal dimension.
    
    \item Flatten: this layer reshapes input data into a one dimensional array, enabling compatibility between Bayesian convolutional layers and Bayesian dense layers. 
    
    \item Dense Reparameterization: this layer implements a reparameterization estimator for Bayesian variational inference.
    It implements a stochastic forward pass via sampling from the kernel and bias distributions. This approach improves the robustness of the model, allowing uncertainty estimation in parameter values and supporting probabilistic modeling in deep learning.

    \item Distribution Lambda: this layer is responsible for producing the final results given the inputs and the learned weights from the previous layers. The output layer consists of two neurons representing the mean, $\hat{y}$ and variance, $\hat{\sigma}^2$, in order to quantify the expected value and its associated uncertainty. To ensure a positive variance, the neuron is activated using an exponential function.
\end{itemize}

BCNN combines feature extraction capabilities of classical CNN models with the uncertainty quantification of Bayesian theory. The proposed architecture is built using the Bayesian layers of \texttt{TensorFlow Probability} in Python \cite{dillon2017}.

\subsubsection{Training for Diversity}




Model diversity is a key concept for effective ensemble models \cite{Nam2021}. Accordingly, in this case, the training set for each battery model is modified to learn different battery aging properties. Historical capacity fading data are used to build aging models for each battery in the dataset.

Namely, using the LOOCV strategy, if $K$ run-to-failure trajectories are available, $K$ diverse BCNN models are built changing the training set in each iteration (cf. Figure~\ref{fig:FlowchartMethodology}). That is, the model is trained on all batteries except one, which is held as a test set. This process is repeated so that each battery serves as a test set exactly once. Thus, all available data are used for training, maximizing the diversity of training scenarios. 

Training the BCNN models through LOOCV strategy, enhances the ability of individual models to generalize across different battery types and manufacturing conditions. 

This stage completes the offline training process, which results in a set of BCNN models:
\begin{equation}
\label{eq:ensemble}
    \mathcal{M} = \{BCNN_1,BCNN_2,\ldots,BCNN_{K}\},
\end{equation}

which are used in the subsequent online inference process to build ensemble models.

\subsection{Online: Stacking of Predictive Distribution}

During the online phase, the proposed stacking of predictive distribution strategy is designed and tested. The proposed approach takes as input individual base models [cf. Eq.~(\ref{eq:ensemble})] and monitored data up to the prediction instant $t$, which is used to forecast the probability density function (PDF) of the capacity at $t+1$, $\hat{y}_{PDF}(t+1)$. The objective of the stacking process is to integrate the predictive distributions of different base models and propagate all the information end-to-end.

For comparison and benchmarking purposes, an alternative stacking approach is also implemented named stacking of point prediction (cf. Subsection \ref{ss:Benchmarking}).

\textbf{Log-Score Weights}

The optimal way to combine a set of Bayesian posterior predictive distributions is by using the logarithmic score \cite{yao2018}. This method maximizes the average log-likelihood of the observed data, which is a proper scoring rule used to evaluate the accuracy of probabilistic forecasts. It measures the accuracy of a forecast and penalizes overconfidence and underconfidence in the predicted probability. The logarithmic score is defined as follows:
\begin{equation}
\label{eq:weights_like}
    \scalebox{0.88}{$\displaystyle \hat{w} = \arg\max\limits_{w} \frac{1}{N} \sum\limits_{i=1}^{N}log\sum\limits_{k=1}^{K} w_k p(y_i \mid y_{-i}, M_k) + \lambda_{reg} \sum\limits_{k=1}^K w_k^2 $}
\end{equation}

where $N$ denotes the total number of data points and $K$ denotes the total number of base models. The leave-one-out predictive distribution for each model, \textit{i.e.} $p(y_i \mid y_{-i}, M_k)$, is used to compute the model's prediction for the data point $i$. To avoid overfitting, a regularization term $\lambda_{reg}$ is added to the likelihood function, penalizing large weights.

\begin{figure*}[!ht]
    \centering
    \subfigure[Voltage variation]{\includegraphics[width=0.31\textwidth]{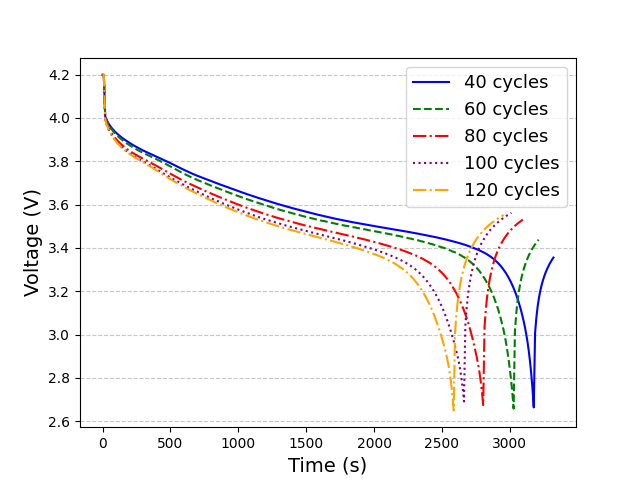}
    \label{subfig:Dis_Volt}} 
    \subfigure[Current variation]{\includegraphics[width=0.31\textwidth]{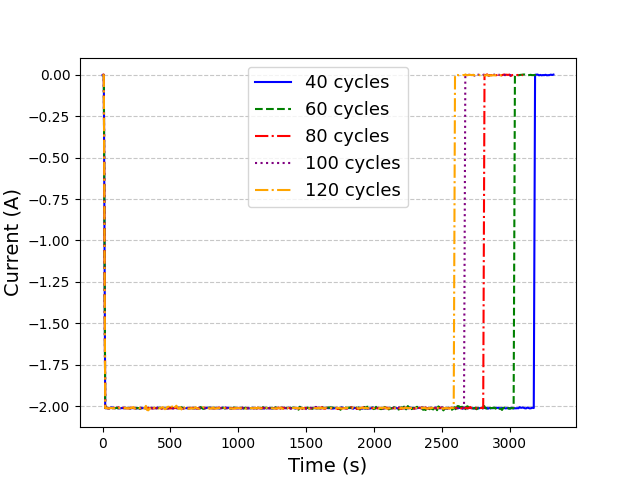}
    \label{subfig:Dis_Current}} 
    \subfigure[Temperature variation]{\includegraphics[width=0.31\textwidth]{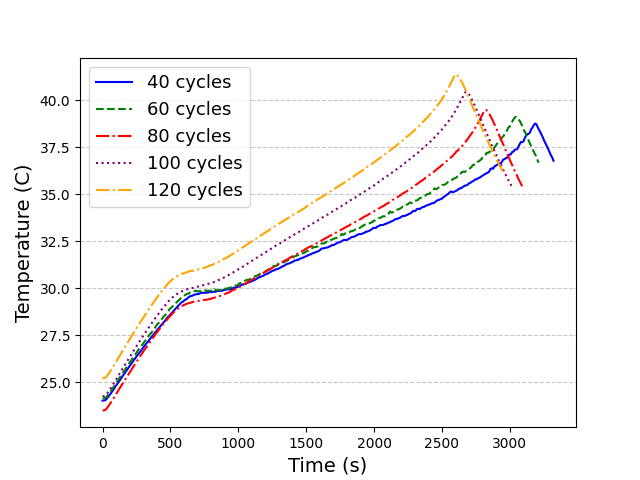}
    \label{subfig:Dis_Temp}} 
    \caption{Feature variations due to an increasing number of discharge cycles in battery \#5.}
    \label{fig:Volt_Current_Temp}
\end{figure*}

\textbf{Stacking}

Stacking is a method to average point estimates from several models \cite{leblanc1996}. In its simplest form, it can be seen as a weighted average method. Through the weighted average, it facilitates the construction of ensembles that incorporate predictions from multiple models. In the proposed framework, the goal of weighted average ensemble is to leverage the predictive capabilities of $K$ pre-trained BCNN models [cf. Eq.~(\ref{eq:ensemble})]. It seeks to mitigate forecasting errors by assigning weights to the linear combination of these models, thereby enhancing the accuracy of predictions.

In the Bayesian framework, stacking extends beyond the limitations of averaging point predictions by combining multiple Bayesian posterior predictive distributions. This approach develops a \textit{ stacking model} that leverages the strengths of various predictive models, enhancing overall predictive accuracy. The stacking of the predictive distribution enables the fusion of uncertainties from various models into a unified predictive framework. This approach improves the accuracy of forecasts and offers a comprehensive evaluation of the uncertainty associated with these forecasts, providing advantages across diverse decision-making scenarios. The fundamental equation governing this process is defined as follows:

\begin{equation}
    \hat{p}(\Tilde{y}|y) = \sum\limits_{k=1}^{K} \hat{w}_k p(\Tilde{y}|y, M_k)
\end{equation}

where $\hat{p}(\Tilde{y}|y)$ represents the aggregate probability estimation based on the ensemble model, $\omega_k$  denotes the weight assigned to the $k$-th component within the ensemble, and $p(\Tilde{y}|y, M_k)$ refers to the probabilistic forecast generated by each base model, denoted as BCNN\textsubscript{k}, given the observed data $y$.

This probabilistic prediction indicates the likelihood of observing the predicted outcome $\Tilde{y}$, dependent on the specific base model employed.

\subsubsection{Forecasting}

Online forecasting is computed for one-step-ahead predictions. In order to forecast battery capacity at instant $t+1$, previous data until the instant $t$ is used, plus an uncertainty factor expressed as noise:

\begin{equation}
\label{eq:persistance_data}
\mathcal{X}(t) = \{V(t), T(t), \epsilon\}
\end{equation}

\noindent where $\{V(t), T(t)\}$ denote the values of voltage and temperature at instant $t$, and $\epsilon$ denotes the Gaussian noise term, $N(0,\sigma)$ with $\sigma=0.1$, that introduces variability in the progression of $X$ over time. 

The one-step-ahead capacity distribution prediction is thus defined as follows:

\begin{equation}
\label{eq:forecast}
    \hat{y}_{PDF}(t+1)=f(\mathcal{X}(t))
\end{equation}

\noindent where $f(.)$, denotes the designed ensemble model, $\hat{y}_{PDF}(t+1)$ is the distribution of the capacity estimate at $t+1$.

It is possible to perform SOH predictions for longer prediction horizons through a recursive forecasting strategy. However, due to the accumulation of individual forecasting errors, this approach may lead to decrease long-term forecasting performance. Long-term SOH forecasting activities are left open for future work.



This approach allows the model to learn continuously and adapt to changing conditions. Online forecasting is particularly beneficial in environments that require immediate decision making based on the latest available data.


\section{Case study}
\label{sec:Case_study}

\subsection{Dataset description}


The effectiveness of the proposed method has been tested using a battery dataset from the NASA Ames Prognostics Center of Excellence \cite{saha2007}. 

A subset of available battery data has been selected, focusing on batteries \#5, \#6, \#7 and \#18. Each battery is operated under various conditions including charging, discharging, and impedance analysis. Throughout the charge and discharge cycles, temperature, current, and voltage were meticulously recorded. During charging, a constant current mode at 1.5 A was maintained until the voltage reached 4.2 V, followed by a switch to constant voltage mode until the current dropped to 20 mA. Discharge cycles involved a constant load mode at 2 A until the voltage levels reached 2.7 V, 2.5 V, 2.2 V and 2.5 V for batteries \#5, \#6, \#7 and \#18, respectively. The experiment ended once the battery capacity decreased by 30\%. These batteries had a maximum capacity of 2Ah with an end-of-life capacity set at 1.4Ah.

\begin{figure}[!ht]
    \centering
    \includegraphics[width=0.95\linewidth]{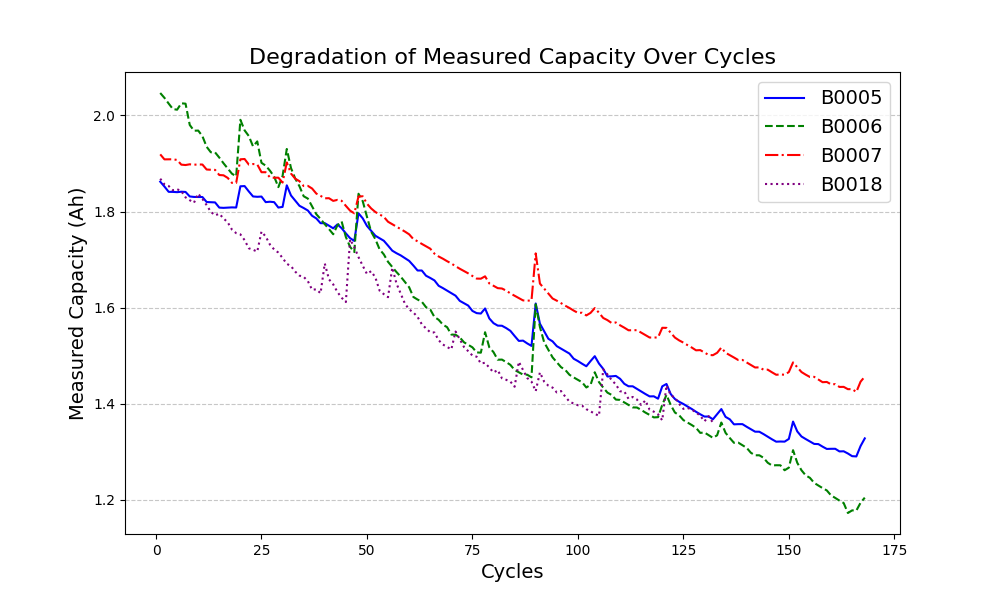}
    \caption{Capacity degradation data of Li-ion batteries.}
    \label{fig:Capacity_fade}
\end{figure}

Figures~\ref{subfig:Dis_Volt},~\ref{subfig:Dis_Current} and~\ref{subfig:Dis_Temp} show the evolution of voltage, current (constant), and temperature measurements with the increment of discharge cycles for the battery \#5. Figure~\ref{fig:Capacity_fade} shows variations in capacity degradation rates for identical batteries. This is an indicator of uncertainty inherent in the manufacturing process, which affects SOH estimates.








\subsection{BCNN structure and hyperparameters}

The design of the base BCNN model structure is developed through experimentation. The BCNN architecture for SOH forecasting is detailed in Table~\ref{table:model_architecture}, where 'None' is indicative of the batch size. The input for the model comprises 371 data points per discharge cycle, with each point aggregating 3 features: voltage, temperature, and time.  

The proposed structure encompasses a total of 1300 trainable parameters, designed to extract features from battery discharge cycle data for forecasting purposes. Figure~\ref{fig:BCNN} details the convolutional layer hyperparameters, which includes 16 kernels, each with a dimension of 3, adopting a Laplace distribution for the prior and employing a ReLU activation function. In addition, the model incorporates Bayesian dense layers with 16 units, Adam optimizer, a learning rate of 0.01, and Evidence Lower Bound (ELBO) as its loss function \cite{Zhang2022}.

\begin{table}[!ht]
    \centering
    \caption{BCNN model architecture}
    \label{table:model_architecture}
    \setlength{\tabcolsep}{3pt}
    \begin{tabular}{@{}llll@{}}
        \toprule
        Layer & Description & Output Shape & \# Param. \\ 
        \midrule
        - & Input & (None, 371, 4) & 0 \\
        1 & Conv.1D Reparameter.     & (None, 369, 16) & 416 \\
        2 & Conv.1D Reparameter.     & (None, 368, 8)  & 528 \\
        3 & Global Average Pooling   & (None, 8)       & 0   \\
        4 & Flatten                  & (None, 8)       & 0   \\
        5 & Dense Reparameter.       & (None, 16)      & 288 \\
        6 & Dense Reparameter.       & (None, 2)       & 68  \\
        7 & Distribution Lambda      & (None,1),(None,1)      & 0  \\ 
        \midrule
        \multicolumn{3}{r}{Total params: 1300 (5.08 KB)} \\
        \bottomrule
    \end{tabular}
    \vspace{-4mm}
\end{table}

\subsection{Benchmarking}
\label{ss:Benchmarking}

In order to compare the designed stacking approach with alternative stacking strategies, another stacking approach has been designed using point prediction information instead of the full distribution.

\textbf{Stacking of Point Prediction}

An effective method for determining the weight of each model in the stacking process is by minimizing the leave-one-out mean squared error with a $L_2$ regularization term, $\lambda_{reg}$. The purpose of this term is to penalize large weights, thus preventing overfitting and balancing individual model contributions. The weights are obtained through the following optimization problem:

\begin{equation}
\label{eq:weights_mse}
    \scalebox{0.92}{$\displaystyle \hat{w} = \arg\min\limits_w \sum\limits_{i=1}^n \left( y_i - \sum\limits_{k=1}^K w_k \hat{f}^{(-i)}_K(x_i) \right)^2 + \lambda_{reg} \sum\limits_{k=1}^K w_k^2 $}
\end{equation}

\noindent where $\hat{f}^{(-i)}_K(x_i)$ represents the predicted value of the $k$-th model, when the $i$-th observation is left out of the training set. The regularization parameter, $\lambda_{reg}$, controls the strength of the regularization applied. To ensure a feasible solution, the weights are restricted to $w_k \ge 0$ and $\sum\limits_{k=1}^{K} w_k = 1$. 



Accordingly, the stacking of point prediction approach is defined as follows:

\begin{equation}
    \hat{y} = \sum\limits_{k=1}^{K} \hat{w}_k f_k(x|\theta_k)
\end{equation}

where $\hat{y}$ represents the prediction of the ensemble for the test battery capacity, $\hat{w}_k$ denotes the weight assigned to the $k$-th battery base model, and $f_k(x|\theta_k)$ is the prediction made by the corresponding base model (BCNN\textsubscript{k}).

\subsection{Evaluation criteria}

The accuracy of the regression is measured by Mean Squared Error, while Negative Log Likelihood assesses model performance by quantifying prediction probabilities. Finally, The correctness of probability predictions is assessed through the CRPS.\\

\noindent \textbf{Mean Square Error} (MSE)   
is a  metric for measuring the quality of an estimator. It is a measure of the average squared differences between the estimated values and what is estimated. MSE is calculated by taking the average of the square of the differences between the predicted values and the actual values~\cite{hodson2022}.

\begin{equation}
\label{eq:mse}
    \text{MSE} = \frac{1}{n} \sum_{i=1}^{n} (Y_i - \hat{Y}_i)^2
\end{equation}

where, $n$ represents the number of observations, $Y_i$ denotes the actual value for the $i$th observation, and $\hat{Y}_i$ signifies the predicted value for the $i$th observation.

\noindent \textbf{Coefficient of Determination} ($\text{R}^2$)   
is a metric used to assess the goodness of fit of the model. It provides a measure of how well the observed outcomes are replicated by the model, based on the proportion of total variation of outcomes explained by the model~\cite{barrett1974}. 

\begin{equation}
\label{eq:r2}
R^2 = 1 - \frac{\sum\limits_{i=1}^{n} (Y_i - \hat{Y}i)^2}{\sum\limits_{i=1}^{n} (Y_i - \bar{Y})^2}
\end{equation}

\noindent where, $n$ is the number of observations, $Y_i$ is the actual value, $\hat{Y}_i$ the predicted value for the $i$-th observation and $\bar{Y}$ the mean of $Y$. $R^2$ of 1 implies perfect model predictions, while 0 means no explained variability.

\noindent \textbf{Continuous Ranked Probability Score} (CRPS) can be formally expressed as a quadratic measure of discrepancy between the predicted Cumulative Distribution Function (CDF), $F(\cdot)$, and the observed empirical CDF for a given scalar observation $y$~\cite{zamo2018}:

\begin{equation}
\label{eq:crps}
    CRPS(F,y) = \int (F(x) -\mathds{I}(x\geq y_i))^2 dx,
\end{equation}

\noindent where $\mathds{I}(x\geq y_i)$ is the indicator function, which models the empirical CDF. 

To obtain a single score value from Eq.~(\ref{eq:crps}), a weighted average is calculated for each individual observation of the test set~\cite{Gneiting2005}:

\begin{equation}
\label{eq:crps_avg}
     CRPS = \frac{1}{N} \sum_{i=1}^{N} CRPS(F_i,y_i)
\end{equation}

\noindent where $N$ denotes the total number of predictions.\\






\noindent \textbf{Negative Log Likelihood} (NLL)  
metric assesses probabilistic models by using the likelihood concept, which indicates how likely the observed data is given model parameters \cite{bosman2000}. Likelihood ($\mathcal{L}$) is the product of each observation's probability density function (PDF), expressed mathematically as 

\begin{equation}
\label{eq:likelihood}
     \mathcal{L}(\theta \mid X) = \prod_{i=1}^{N} f(x_i | \theta)
\end{equation}

where $\theta$ denotes model parameters and $X$ includes $N$ data points. NLL is preferred for optimization since minimizing NLL is equivalent to maximizing the log-likelihood, facilitating the discovery of model parameters that best explain the observed data, represented by 

\begin{equation}
\label{eq:nll}
     -\log \mathcal{L}(\theta \mid X) = -\sum_{i=1}^{n} \log f(x_i \mid \theta)
\end{equation}

\noindent \textbf{Calibration} refers to the statistical consistency between the predictive distributions and the actual observations. It represents a joint property of forecasts and empirical data~\cite{jung2022}. Namely, it is stated that the model is calibrated if~\cite{kuleshov2018}:

    \begin{equation}
        \label{eq:calibration}
        \frac{\sum_{t=1}^{T} \mathds{I}\{y_t \leq F_{t}^{-1}(p)\} }{T} \rightarrow p \text{ for all } p \in [0,1]
    \end{equation}

\noindent In this expression, $T$ refers to the total number of data points, while the indicator function $\mathds{I}\{y_t \leq F_{t}^{-1}(p)\}$ takes a value of 1 when the condition $y_t \leq F_{t}^{-1}(p)$ is true, and 0 otherwise. Given this condition, $y_t$ express the observed outcome at time $t$, and $F_{t}^{-1}(p)$ is the inverse of the CDF for the forecast, evaluated at probability $p$. Therefore, the condition represents the threshold below which a random sample from the distribution would occur with a probability $p$.\\


\noindent \textbf{Sharpness} means that the confidence intervals should be optimized for minimal width around a singular value. That is, the goal is to reduce the variance, denoted as $var(F_n)$, of the random variable characterized by the cumulative distribution function $F_n$~\cite{kuleshov2018, Tran_2020}:

    \begin{equation}
        \label{eq:sharpness}
        sha = \sqrt{\frac{1}{N} \sum_{n=1}^{N} var(F_n)}
    \end{equation}

\begin{table*}[!ht]
    \centering
    \caption{Comparison of different ensemble strategies for different batteries used as test.}
    \setlength{\tabcolsep}{1.5pt}
    \begin{tabular}{lcccccccccccl}
        \toprule
         & \multicolumn{4}{c}{Baseline Model} & \multicolumn{4}{c}{Benchmarking Ensemble} & \multicolumn{4}{c}{Proposed Ensemble}\\
         \cmidrule(lr){2-5} \cmidrule(lr){6-9} \cmidrule(lr){10-13}
         & $MSE(\downarrow)$ & $R^2(\uparrow)$ & $NLL(\downarrow)$ & $CRPS(\downarrow)$ & $MSE(\downarrow)$ & $R^2(\uparrow)$ & $NLL(\downarrow)$ & $CRPS(\downarrow)$ & $MSE(\downarrow)$ & $R^2(\uparrow)$ & $NLL(\downarrow)$ & $CRPS(\downarrow)$ \\
         \midrule
        B0005& 0.0007&0.9732 & 2.3397& 0.0183& \textbf{0.0002}& \textbf{0.9901} & -1.9523& 0.0145& 0.0003& 0.9886 &\textbf{ -2.1001}& \textbf{0.0131}\\
        B0006 & 0.0013 &0.9636 & 8.0947 & 0.0213 & \textbf{0.0009} & \textbf{0.9753} & -1.8222 & 0.0183 & 0.0009 & 0.9741 & \textbf{-1.9358} & \textbf{0.0178}\\
        B0007 & 0.0005 & 0.9696 & -0.0409 & 0.0149 & \textbf{0.0003} & \textbf{0.9814}& \textbf{-1.9755} &\textbf{ 0.0145} & \textbf{0.0004} & \textbf{0.9763}& \textbf{-1.9769} & \textbf{0.0145 }\\
        B0018 & 0.0013 & 0.8943 & 9.0342 & 0.0223  & \textbf{0.0010}  & \textbf{0.9183}& \textbf{-1.9478}  & \textbf{0.0174} & \textbf{0.0010} & \textbf{0.9141} & \textbf{-1.9312} & \textbf{0.0178} \\
        \bottomrule
    \end{tabular}
    \label{table:results}
\end{table*}

\section{Results}
\label{sec:Results}

To evaluate the proposed approach, firstly, different ensemble strategies are compared to evaluate their strengths and identify the most suitable approach. Subsequently, a sensitivity analysis is developed with respect to the contribution of individual base-models to the overall ensemble.

\subsection{Probabilistic Ensemble Strategies}

This section focuses on the comparison between (i) the baseline model, \textit{i.e.} BCNN model trained with all available data, (ii) ensemble of point prediction and (iii) proposed ensemble method (cf. Figure~\ref{fig:FlowchartMethodology}) to further evaluate the improvement of ensemble strategies over baseline model.

Table~\ref{table:results} presents a comparative analysis in terms of accuracy and probabilistic metrics. This comparison highlights that, for different test scenarios, the ensemble methodologies enhance the performance of the baseline model. 

    

A notable observation from the results in Table~\ref{table:results} is the variance between the proposed ensemble approach (cf. Figure~\ref{fig:FlowchartMethodology}) and the benchmarking ensemble model (cf. Subsection \ref{ss:Benchmarking}) in specific scenarios. For batteries \#5 and \#6, the proposed approach exhibited superior outcomes, particularly in probabilistic metrics (NLL and CRPS). This suggests that within a Bayesian framework, prioritizing likelihood maximization, leads to accurately modelling uncertainty, and therefore, it is more advantageous than focusing on MSE minimization (as in Subsection \ref{ss:Benchmarking}). 

The model optimization criterion has a direct impact on the performance of the tested methods and on the effectiveness of the ensemble approach. However, for batteries \#7 and \#18, no significant differences were observed between the tested ensemble approaches, which indicates that the results are associated to the prior models. That is, it is possible that the same prior model minimizes the MSE and maximizes the likelihood at the same time.

\begin{figure}[!ht]
    \centering
    
    \subfigure[Stacked point prediction method (cf. Subsection \ref{ss:Benchmarking}) ]{\includegraphics[width=0.45\textwidth]{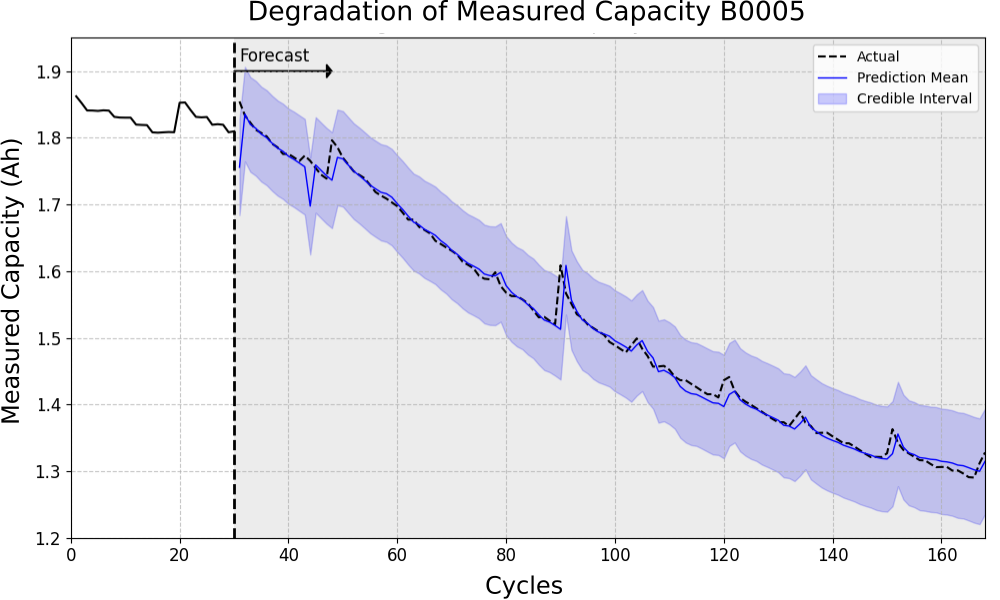}\label{subfig:B0005_Ensemble_sapp}} 
    \subfigure[Stacked predictive distribution method (cf. Figure~\ref{fig:FlowchartMethodology})]{\includegraphics[width=0.45\textwidth]{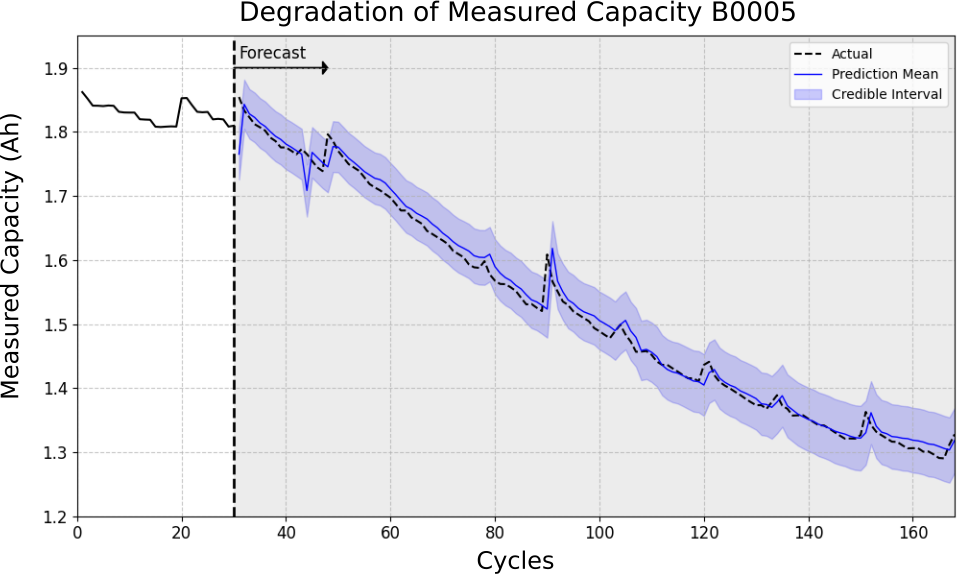}\label{subfig:B00005_Ensemble_sabpd}} 
    \subfigure[Baseline model]{\includegraphics[width=0.45\textwidth]{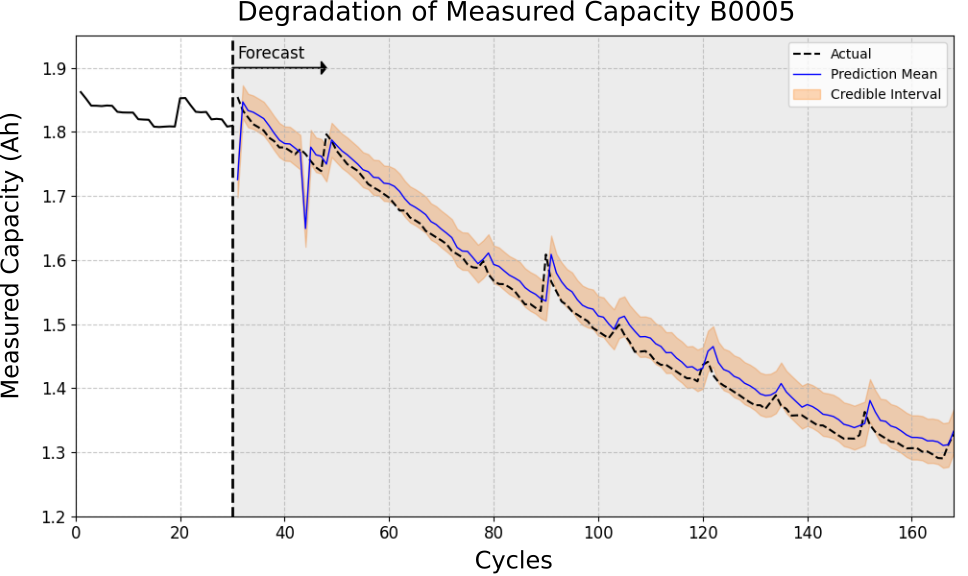}\label{subfig:B0005_No_Ensemble}} 
    \caption{Battery capacity degradation forecasting results.}
    \label{fig:sapp_vs_sabpd_vs_noEnsemble}
\end{figure}

Figure~\ref{subfig:B0005_Ensemble_sapp} shows the comparison between the ensemble model generated by stacking point predictions (cf. Subsection \ref{ss:Benchmarking}), Figure~\ref{subfig:B00005_Ensemble_sabpd} shows the ensemble model generated through stacking of predictive distributions (cf. Figure~\ref{fig:FlowchartMethodology}), and Figure~\ref{subfig:B0005_No_Ensemble} shows the individual BCNN trained with the entire dataset, e.g. for the battery \#5, train with batteries \#6, \#7, and \#18, and test with \#5. 

\begin{figure*}[!ht]
    \centering
    \subfigure[calibration and sharpness for the benchmarking ensemble model]{\includegraphics[width=0.49\textwidth]{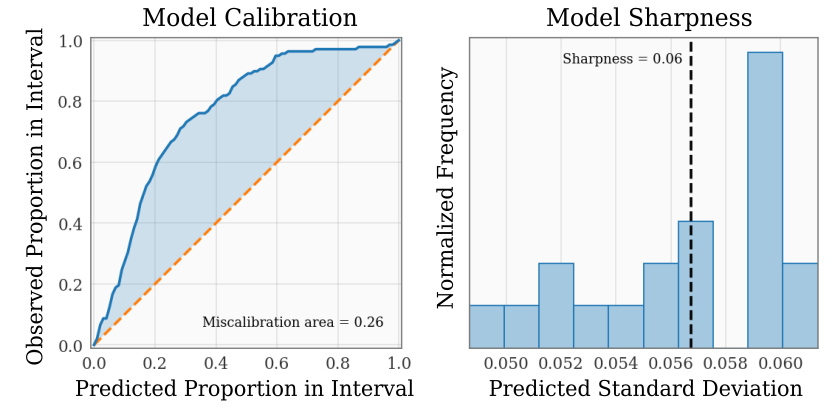}\label{subfig:B0005_Ensemble_mse_cal_sharp}} 
    \subfigure[calibration and sharpness for the proposed ensemble model]{\includegraphics[width=0.49\textwidth]{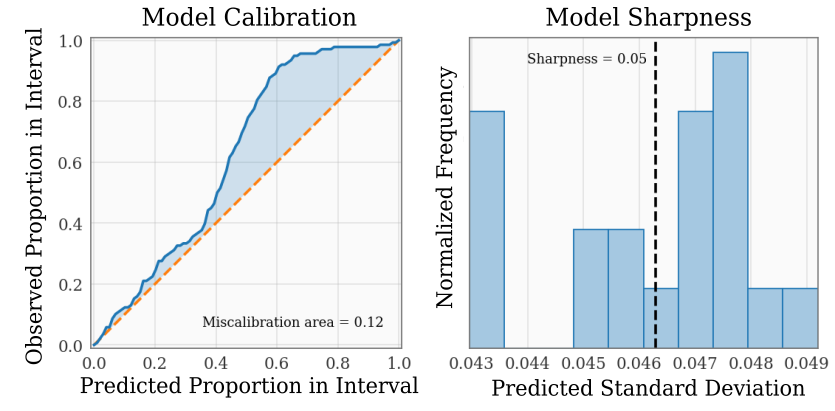}
    \label{subfig:B0005_Ensemble_Like_cal_sharp}} 
    \caption{Evaluation of calibration and sharpness for battery \#5.}
    \label{fig:B0005_Ensemble_cal_sharp}
\end{figure*}

It is observed that the ensemble models enhance the performance of baseline model in terms of accuracy and uncertainty quantification. This is indicated by the positioning of the ground truth (dashed lines) at the limit of the lower boundary in Figure~\ref{subfig:B0005_No_Ensemble}, which means that the uncertainty does not accurately cover the observed values. That is, the uncertainty bounds are not well-calibrated, compromising the model's ability to accurately represent the underlying variability in the data and in the model compared to ensemble strategies.


Figure \ref{subfig:B0005_Ensemble_sapp} shows an improvement in the prediction accuracy. However, it simultaneously introduces a higher level of uncertainty compared to the proposed ensemble method in Figure \ref{subfig:B00005_Ensemble_sabpd}. This is reflected in the NLL and CRPS metrics, where the stacking of the predictive distribution demonstrates superior performance (cf. Table~\ref{table:results}). Such probabilistic metrics indicate that the model parameters make the observed data more probable, indicating a good fit to the observed data.

The evaluation of the shape of the PDF is a crucial aspect of uncertainty quantification. Accordingly, the calibration and the sharpness assessment of PDFs is performed through a python toolbox for predictive uncertainty quantification~\cite{chung2021uncertainty}. Figure~\ref{fig:B0005_Ensemble_cal_sharp} shows the calibration and sharpness of the analysed ensemble methods designed for probabilistic forecasting for the battery \#5.

The calibration plot for the point-prediction ensemble model [cf. Figure~\ref{subfig:B0005_Ensemble_mse_cal_sharp}] reveals a miscalibration area of 0.26, indicating a gap between predicted probabilities and actual outcomes, generally overestimating event probabilities. On the contrary, the proposed ensemble model [cf. Figure~\ref{subfig:B0005_Ensemble_Like_cal_sharp}] shows better calibration with a miscalibration area of 0.12, aligning closer to the ideal, especially in midrange probabilities. 

In terms of sharpness, the predictions of the point-prediction based ensemble model have a mean sharpness value of 0.06 and are right-skewed, reflecting higher uncertainty. However, the proposed ensemble model has a mean sharpness value of 0.05, with a slightly left-skewed distribution, indicating more predictions with lower uncertainty and greater confidence.

\subsection{Sensitivity of the Ensemble Strategy with Base-Models}
\label{ss:Sensitivity}

To evaluate the contribution of each individual BCNN model to the ensemble approach, a sensitivity assessment has been performed. Namely, the performance of the different leave-one-out iterations has been evaluated, sequentially training with different battery datasets and testing with the leave-out battery dataset. This has been compared with the proposed ensemble approach results to identify individual contributions from different models. Table~\ref{table:result_Ensemble_vs_BCNN} displays the obtained results.

\begin{table}[!ht]
    \begin{threeparttable}[b]
    \centering
    \caption{Performance evaluation of BCNN models and the ensemble approach.}
    \setlength{\tabcolsep}{3.5pt}
    \begin{tabular}{lcccccl}
       \toprule
           Test\tnote{1} & Model & \textit{MSE} ($\downarrow$) & \textit{$\text{R}^2$} ($\uparrow$) & \textit{NLL} ($\downarrow$) & \textit{CRPS} ($\downarrow$) \\
         \midrule
         \multirow{4}{*}{\#5} & BCNN [\#6,\#7]\tnote{2} & 0.0005 & 0.9802  & -1.0707 & 0.0135 \\
         & BCNN [\#6,\#18]  & 0.0244 & 0.1016  & 19.4417  & 0.1411\\
        & BCNN [\#7,\#18]  & 0.0006 & 0.9795  & -2.0774 & 0.0132  \\
        & Ensemble& \textbf{0.0003}& \textbf{0.9886} & \textbf{-2.1001}  & \textbf{0.0131} \\[5pt]
        
        \multirow{4}{*}{\#6} & BCNN [\#5,\#7] & 0.0011 & 0.9695  & 3.7012  & 0.0197  \\
        & BCNN [\#5,\#18]  &  0.0147 & 0.5861  & 0.5852  & 0.0849 \\
        & BCNN [\#7,\#18] & 0.0018 & 0.9491  & -0.7498 & 0.0252 \\
        & Ensemble & \textbf{0.0009}& \textbf{0.9741} & \textbf{-1.9358}  & \textbf{0.0178} \\[5pt]

        \multirow{4}{*}{\#7} & BCNN [\#5,\#6]  & 0.0008 & 0.9543 & -1.5462 & 0.0166 \\
        & BCNN [\#5,\#18]&  0.004  & 0.7704 & 2.1996  & 0.0326 \\
        & BCNN [\#6,\#18] & 0.0026 & 0.854  & -1.5735 & 0.0286 \\
        & Ensemble & \textbf{0.0004}& \textbf{0.9763} & \textbf{-1.9769}  & \textbf{0.0145} \\[5pt]

        \multirow{4}{*}{\#18} & BCNN [\#5,\#6] & 0.0091 & 0.2534 & 14.708 & 0.0833 \\
        & BCNN [\#5,\#7] & 0.0041 & 0.6663 & 1.5441 & 0.0459 \\
        & BCNN [\#6,\#7] & 0.0013 & 0.8929 & 1.8299 & 0.0213 \\
        & Ensemble & \textbf{0.0010}& \textbf{0.9141} & \textbf{-1.9312}  & \textbf{0.0178} \\
        \bottomrule
    \end{tabular}
    \begin{tablenotes}
       \item [1] Battery identifier used for testing.
       \item [2] BCNN [\#A,\#B]: BCNN trained with batteries \#A and \#B.
     \end{tablenotes}
    \label{table:result_Ensemble_vs_BCNN}
    \end{threeparttable}
\end{table}

The ensemble BCNN model demonstrates significantly higher accuracy and predictive power than individual BCNN models, as evidenced by its superior performance across multiple metrics. It achieves the lowest MSE in every testing battery, indicating more precise predictions, and the highest $\text{R}^2$ score, showing its ability to explain a greater proportion of variance. 

 \begin{figure*}[!ht]
     \centering
     \subfigure[Ensemble forecast showing the combined prediction from all models]{\includegraphics[width=0.495\textwidth]{Figures/Capacity_fade_Ensemble_B0005_SABPD.png}\label{subfig:B0005_Ensemble}} 
     \subfigure[Forecast from the first component model of the ensemble]{\includegraphics[width=0.495\textwidth]{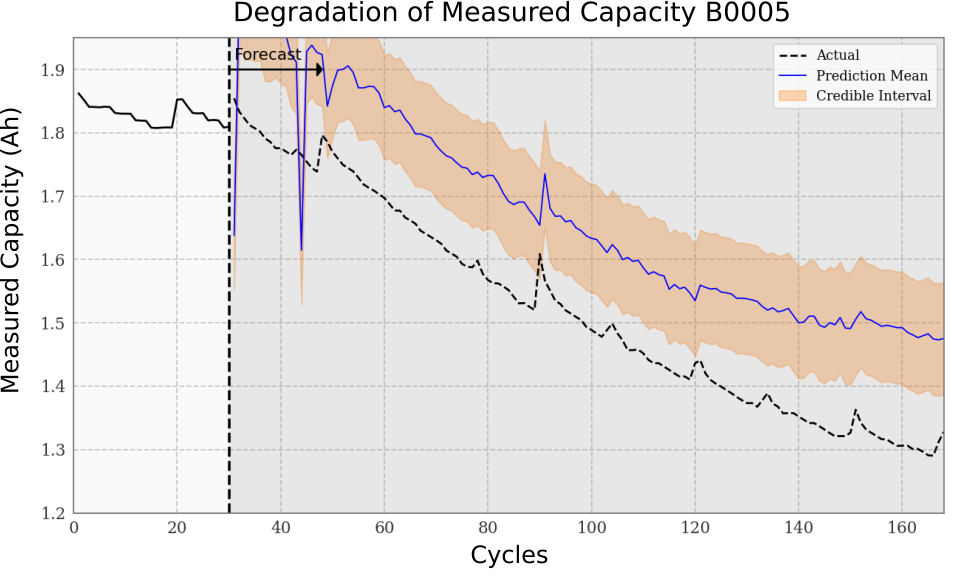}\label{subfig:B0005_model0}} 
     \subfigure[Forecast from the second component model of the ensemble]{\includegraphics[width=0.495\textwidth]{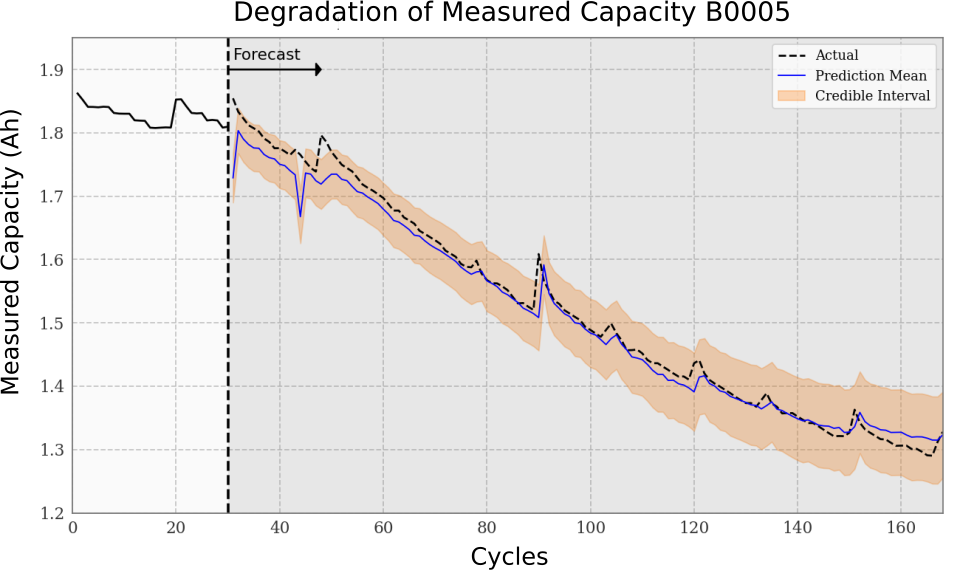}\label{subfig:B0005_model1}} 
     \subfigure[Forecast from the third component model of the ensemble]{\includegraphics[width=0.495\textwidth]{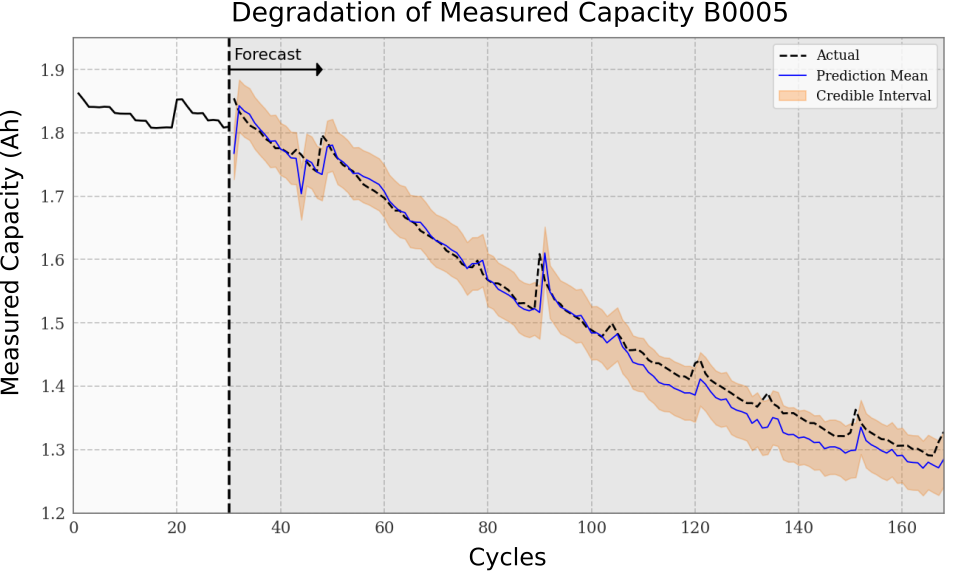}\label{subfig:B0005_model2}} 
     \caption{Capacity fade forecasting for battery \#5 employing an ensemble of BCNN models.}
     \label{fig:ensemble_models}
 \end{figure*}

The ensemble model also shows a notable improvement in the NLL metric, suggesting a more reliable uncertainty estimation. Additionally, by achieving the lowest CRPS, it emphasizes its proficiency in probabilistic forecasting and precise uncertainty quantification. Overall, the ensemble method outperforms individual models, highlighting its effectiveness in contexts that require high accuracy and reliability.

Figure~\ref{fig:ensemble_models} presents the forecasts generated by individual models for battery \#5 (cf. Table~\ref{table:result_Ensemble_vs_BCNN}). Figures \ref{subfig:B0005_model0}-\ref{subfig:B0005_model2}, show individual models and Figure~\ref{subfig:B0005_Ensemble} shows the combined forecast of the ensemble model.

It can be seen that the ensemble effectively combines the characteristics of models 2 and 3, thereby improving the overall performance of the final forecast of the ensemble.

\section{Discussion}
\label{sec:Discussion}

The proposed research work demonstrates that the stacking of predictive distributions based on a Bayesian framework improves the accuracy and robustness of predictions compared with stacking of point predictions. Furthermore, it has been observed that the use of an ensemble of BCNN models improves the modeling of uncertainty when compared to relying on a single BCNN model (baseline). However, before drawing definitive conclusions about the application of the proposed solution in real-world applications, further work is necessary testing the robustness, scalability, and sensitivity with respect to noise.




\textit{Robustness}

Credible intervals reflect the uncertainty associated with the data and the model (cf. Figure~\ref{fig:sapp_vs_sabpd_vs_noEnsemble}). The robustness of the proposed approach is therefore directly dependent on model and data uncertainty. The reduction of credible intervals align with the objective of increasing robustness. To this end, increasing the number of observations would reduce the uncertainty attributed to the model, which results in more precise credible intervals. Additionally, employing priors like maximum entropy priors or weakly informative priors may further tighten credible intervals, thereby improving the reliability of the model predictions.

\textit{Scalability}

To analyze larger fleets of batteries, instead of using leave-one-out methodologies, it may be more appropriate to develop generalized training methodologies. In this direction, one approach would be to cluster batteries that exhibit similar operation and degradation conditions. This strategy would enable capturing data diversity, which is a key property for ensemble strategies. Alternatively, a hierarchical modelling strategy may be adopted. This method involves a global model for overall battery behavior, supplemented by smaller models for specific groups, enabling precise adaptations without the need for separate models per battery. This strategy ensures scalability and flexibility in handling various battery operation and degradation conditions efficiently.

\textit{Noise Sensitivity}

\begin{figure}[!htb]
    \centering
    \includegraphics[width=0.99\linewidth]{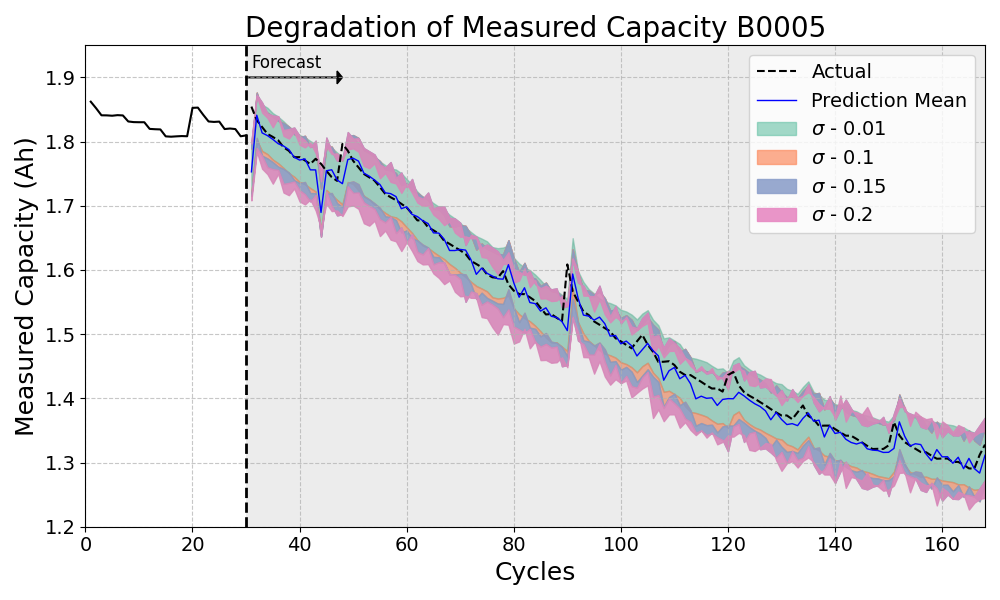}
    \caption{Impact of Gaussian Noise on Predictive Modeling of Battery Capacity Degradation.}
    \label{fig:noise_sens}
\end{figure}

The proposed approach assumes a Gaussian noise to model the variability of the modeled process and measurements [cf. Eq.~(\ref{eq:persistance_data})]. To analyze the impact of Gaussian noise levels on prediction results, a sensitivity analysis has been performed. Figure~\ref{fig:noise_sens} shows the obtained results.

 Obtained results indicate that, when testing data diverges from training data, the epistemic uncertainty increases. The increase in Gaussian noise causes a greater deviation, and therefore, there is a significant rise in epistemic uncertainty. Analysing the model's behaviour in the presence of different types of uncertainty is crucial to evaluate the robustness of the model and determine if additional training stages are needed to enhance its reliability. Consequently, this research adopts a noise level of 0.1 as a trade-off decision between prediction accuracy and uncertainty.

\textit{Application Limits}

Some of the adopted practices may limit the applicability of the proposed framework in real-world applications. The experimental setup, conducted in a controlled environment with specified load conditions, may not entirely replicate the diverse sources of uncertainty present in real-world applications. Such controlled conditions could potentially skew the understanding of uncertainty due to environmental and operational variabilities. Consequently, the predictive performance observed in this study may differ under less predictable conditions. In this direction, for controlled operation environments, the complexity of the proposed approach may be reduced. However, the proposed methodology complexity is designed to capture a wide range of uncertainties found in real operating systems.







\section{Conclusion and future work}
\label{sec:Conclusions}

Batteries are key components in power and energy systems and ensuring a robust and reliable remaining useful life (RUL) prediction of batteries is crucial to develop accurate monitoring strategies, and build cost-effective solutions.

In this context, battery RUL prediction models generally focus on individual prediction models. They may be able to capture uncertainty associated with the battery ageing process, but the uncertainty modelling and capturing ability is also limited to the individual model. This research presents a probabilistic ensemble prognostics approach which combines Bayesian Convolutional Neural Network (BCNN) models in a probabilistic stacking strategy. The proposed framework leverages the probabilistic predictive information of individual BCNN models, which are integrated through a probabilistic stacking approach that calibrates between accuracy and robustness of probabilistic predictions.

The proposed approach has been tested on NASA's battery dataset. Obtained results show that the proposed  probabilistic stacking approach improves accuracy and uncertainty of predictions with respect to other ensemble strategies and individual BCNN models.

This research study contributes towards understanding and predicting the capacity fade in Li-ion batteries. Namely, it highlights the role of probabilistic approaches and ensemble methods in modelling the uncertainties inherent in battery manufacturing and operation. 

Looking forward, there are different opportunities to expand the scope and applicability of this work. On the one hand, the use of a larger battery dataset, which includes diverse environmental and operational conditions, would allow for a more comprehensive understanding of capacity fade across various scenarios. On the other hand, it may be possible to perform a more exhaustive comparative analysis of different fusion strategies, including Bayesian Model Averaging, Pseudo Bayesian Model Averaging, or Mixture Models. This comparative will provide further insights into the optimal approaches for integrating predictive models in the context of battery life prediction, enhancing both the accuracy and reliability of capacity fade forecasts.


\section*{Acknowledgenemts}

This publication is part of the research projects KK-2023-00041, IT1451-22 and IT1676-22 funded by the Basque Government. J. I. Aizpurua is funded by Juan de la Cierva Incorporacion Fellowship, Spanish State Research Agency (grant No. IJC2019-039183-I).

\bibliographystyle{apacite}

\begin{thebibliography}{}

\bibitem [\protect \citeauthoryear {%
Abdar%
\ \protect \BOthers {.}}{%
Abdar%
\ \protect \BOthers {.}}{%
{\protect \APACyear {2021}}%
}]{%
abdar2021}
\APACinsertmetastar {%
abdar2021}%
\begin{APACrefauthors}%
Abdar, M.%
, Pourpanah, F.%
, Hussain, S.%
, Rezazadegan, D.%
, Liu, L.%
, Ghavamzadeh, M.%
\BDBL {}Nahavandi, S.%
\end{APACrefauthors}%
\unskip\
\newblock
\APACrefYearMonthDay{2021}{{\APACmonth{12}}}{}.
\newblock
{\BBOQ}\APACrefatitle {A Review of Uncertainty Quantification in Deep Learning: {{Techniques}}, Applications and Challenges} {A review of uncertainty quantification in deep learning: {{Techniques}}, applications and challenges}.{\BBCQ}
\newblock
\APACjournalVolNumPages{Information Fusion}{76}{}{243--297}.
\newblock
\begin{APACrefDOI} \doi{10.1016/j.inffus.2021.05.008} \end{APACrefDOI}
\PrintBackRefs{\CurrentBib}

\bibitem [\protect \citeauthoryear {%
Bai%
\ \BBA {} Chandra%
}{%
Bai%
\ \BBA {} Chandra%
}{%
{\protect \APACyear {2023}}%
}]{%
bai2023a}
\APACinsertmetastar {%
bai2023a}%
\begin{APACrefauthors}%
Bai, G.%
\BCBT {}\ \BBA {} Chandra, R.%
\end{APACrefauthors}%
\unskip\
\newblock
\APACrefYearMonthDay{2023}{{\APACmonth{11}}}{}.
\newblock
{\BBOQ}\APACrefatitle {Gradient Boosting {{Bayesian}} Neural Networks via {{Langevin MCMC}}} {Gradient boosting {{Bayesian}} neural networks via {{Langevin MCMC}}}.{\BBCQ}
\newblock
\APACjournalVolNumPages{Neurocomputing}{558}{}{126726}.
\newblock
\begin{APACrefDOI} \doi{10.1016/j.neucom.2023.126726} \end{APACrefDOI}
\PrintBackRefs{\CurrentBib}

\bibitem [\protect \citeauthoryear {%
Barrett%
}{%
Barrett%
}{%
{\protect \APACyear {1974}}%
}]{%
barrett1974}
\APACinsertmetastar {%
barrett1974}%
\begin{APACrefauthors}%
Barrett, J\BPBI P.%
\end{APACrefauthors}%
\unskip\
\newblock
\APACrefYearMonthDay{1974}{}{}.
\newblock
{\BBOQ}\APACrefatitle {The coefficient of determination—some limitations} {The coefficient of determination—some limitations}.{\BBCQ}
\newblock
\APACjournalVolNumPages{The American Statistician}{28}{1}{19--20}.
\PrintBackRefs{\CurrentBib}

\bibitem [\protect \citeauthoryear {%
Blundell%
, Cornebise%
, Kavukcuoglu%
\BCBL {}\ \BBA {} Wierstra%
}{%
Blundell%
\ \protect \BOthers {.}}{%
{\protect \APACyear {2015}}%
}]{%
blundell2015}
\APACinsertmetastar {%
blundell2015}%
\begin{APACrefauthors}%
Blundell, C.%
, Cornebise, J.%
, Kavukcuoglu, K.%
\BCBL {}\ \BBA {} Wierstra, D.%
\end{APACrefauthors}%
\unskip\
\newblock
\APACrefYearMonthDay{2015}{}{}.
\newblock
{\BBOQ}\APACrefatitle {Weight Uncertainty in Neural Network} {Weight uncertainty in neural network}.{\BBCQ}
\newblock
\BIn{} F.~Bach\ \BBA {} D.~Blei\ (\BEDS), \APACrefbtitle {International conference on machine learning} {International conference on machine learning}\ (\BVOL~37, \BPGS\ 1613--1622).
\newblock
\APACaddressPublisher{Lille, France}{PMLR}.
\PrintBackRefs{\CurrentBib}

\bibitem [\protect \citeauthoryear {%
Bosman%
\ \BBA {} Thierens%
}{%
Bosman%
\ \BBA {} Thierens%
}{%
{\protect \APACyear {2000}}%
}]{%
bosman2000}
\APACinsertmetastar {%
bosman2000}%
\begin{APACrefauthors}%
Bosman, P\BPBI A.%
\BCBT {}\ \BBA {} Thierens, D.%
\end{APACrefauthors}%
\unskip\
\newblock
\APACrefYearMonthDay{2000}{}{}.
\newblock
{\BBOQ}\APACrefatitle {Negative log-likelihood and statistical hypothesis testing as the basis of model selection in IDEAs} {Negative log-likelihood and statistical hypothesis testing as the basis of model selection in ideas}.{\BBCQ}
\newblock
\BIn{} \APACrefbtitle {Proceedings of the Tenth Dutch--Netherlands Conference on Machine Learning. Tilburg University.} {Proceedings of the tenth dutch--netherlands conference on machine learning. tilburg university.}
\PrintBackRefs{\CurrentBib}

\bibitem [\protect \citeauthoryear {%
Che%
\ \protect \BOthers {.}}{%
Che%
\ \protect \BOthers {.}}{%
{\protect \APACyear {2024}}%
}]{%
che2024}
\APACinsertmetastar {%
che2024}%
\begin{APACrefauthors}%
Che, Y.%
, Zheng, Y.%
, Forest, F\BPBI E.%
, Sui, X.%
, Hu, X.%
\BCBL {}\ \BBA {} Teodorescu, R.%
\end{APACrefauthors}%
\unskip\
\newblock
\APACrefYearMonthDay{2024}{{\APACmonth{01}}}{}.
\newblock
{\BBOQ}\APACrefatitle {Predictive Health Assessment for Lithium-Ion Batteries with Probabilistic Degradation Prediction and Accelerating Aging Detection} {Predictive health assessment for lithium-ion batteries with probabilistic degradation prediction and accelerating aging detection}.{\BBCQ}
\newblock
\APACjournalVolNumPages{Reliability Engineering \& System Safety}{241}{}{109603}.
\newblock
\begin{APACrefDOI} \doi{10.1016/j.ress.2023.109603} \end{APACrefDOI}
\PrintBackRefs{\CurrentBib}

\bibitem [\protect \citeauthoryear {%
Chung%
, Char%
, Guo%
, Schneider%
\BCBL {}\ \BBA {} Neiswanger%
}{%
Chung%
\ \protect \BOthers {.}}{%
{\protect \APACyear {2021}}%
}]{%
chung2021uncertainty}
\APACinsertmetastar {%
chung2021uncertainty}%
\begin{APACrefauthors}%
Chung, Y.%
, Char, I.%
, Guo, H.%
, Schneider, J.%
\BCBL {}\ \BBA {} Neiswanger, W.%
\end{APACrefauthors}%
\unskip\
\newblock
\APACrefYearMonthDay{2021}{}{}.
\newblock
{\BBOQ}\APACrefatitle {Uncertainty Toolbox: an Open-Source Library for Assessing, Visualizing, and Improving Uncertainty Quantification} {Uncertainty toolbox: an open-source library for assessing, visualizing, and improving uncertainty quantification}.{\BBCQ}
\newblock
\APACjournalVolNumPages{arXiv preprint arXiv:2109.10254}{}{}{}.
\PrintBackRefs{\CurrentBib}

\bibitem [\protect \citeauthoryear {%
Cobb%
\ \protect \BOthers {.}}{%
Cobb%
\ \protect \BOthers {.}}{%
{\protect \APACyear {2019}}%
}]{%
cobb2019}
\APACinsertmetastar {%
cobb2019}%
\begin{APACrefauthors}%
Cobb, A\BPBI D.%
, Himes, M\BPBI D.%
, Soboczenski, F.%
, Zorzan, S.%
, O'Beirne, M\BPBI D.%
, Baydin, A\BPBI G.%
\BDBL {}Angerhausen, D.%
\end{APACrefauthors}%
\unskip\
\newblock
\APACrefYearMonthDay{2019}{{\APACmonth{06}}}{}.
\newblock
{\BBOQ}\APACrefatitle {An {{Ensemble}} of {{Bayesian Neural Networks}} for {{Exoplanetary Atmospheric Retrieval}}} {An {{Ensemble}} of {{Bayesian Neural Networks}} for {{Exoplanetary Atmospheric Retrieval}}}.{\BBCQ}
\newblock
\APACjournalVolNumPages{The Astronomical Journal}{158}{1}{33}.
\newblock
\begin{APACrefDOI} \doi{10.3847/1538-3881/ab2390} \end{APACrefDOI}
\PrintBackRefs{\CurrentBib}

\bibitem [\protect \citeauthoryear {%
Dai%
, Pollock%
\BCBL {}\ \BBA {} Roberts%
}{%
Dai%
\ \protect \BOthers {.}}{%
{\protect \APACyear {2023}}%
}]{%
dai2023}
\APACinsertmetastar {%
dai2023}%
\begin{APACrefauthors}%
Dai, H.%
, Pollock, M.%
\BCBL {}\ \BBA {} Roberts, G\BPBI O.%
\end{APACrefauthors}%
\unskip\
\newblock
\APACrefYearMonthDay{2023}{{\APACmonth{02}}}{}.
\newblock
{\BBOQ}\APACrefatitle {Bayesian Fusion: Scalable Unification of Distributed Statistical Analyses} {Bayesian fusion: Scalable unification of distributed statistical analyses}.{\BBCQ}
\newblock
\APACjournalVolNumPages{Journal of the Royal Statistical Society Series B: Statistical Methodology}{85}{1}{84--107}.
\newblock
\begin{APACrefDOI} \doi{10.1093/jrsssb/qkac007} \end{APACrefDOI}
\PrintBackRefs{\CurrentBib}

\bibitem [\protect \citeauthoryear {%
Dillon%
\ \protect \BOthers {.}}{%
Dillon%
\ \protect \BOthers {.}}{%
{\protect \APACyear {2017}}%
}]{%
dillon2017}
\APACinsertmetastar {%
dillon2017}%
\begin{APACrefauthors}%
Dillon, J\BPBI V.%
, Langmore, I.%
, Tran, D.%
, Brevdo, E.%
, Vasudevan, S.%
, Moore, D.%
\BDBL {}Saurous, R\BPBI A.%
\end{APACrefauthors}%
\unskip\
\newblock
\APACrefYearMonthDay{2017}{{\APACmonth{11}}}{}.
\newblock
\APACrefbtitle {{{TensorFlow Distributions}}} {{{TensorFlow Distributions}}}\ (\BNUM\ arXiv:1711.10604).
\newblock
\APACaddressPublisher{}{arXiv}.
\PrintBackRefs{\CurrentBib}

\bibitem [\protect \citeauthoryear {%
Fan%
, Olson%
\BCBL {}\ \BBA {} Evans%
}{%
Fan%
\ \protect \BOthers {.}}{%
{\protect \APACyear {2017}}%
}]{%
fan2017}
\APACinsertmetastar {%
fan2017}%
\begin{APACrefauthors}%
Fan, Y.%
, Olson, R.%
\BCBL {}\ \BBA {} Evans, J\BPBI P.%
\end{APACrefauthors}%
\unskip\
\newblock
\APACrefYearMonthDay{2017}{}{}.
\newblock
{\BBOQ}\APACrefatitle {A Bayesian posterior predictive framework for weighting ensemble regional climate models} {A bayesian posterior predictive framework for weighting ensemble regional climate models}.{\BBCQ}
\newblock
\APACjournalVolNumPages{Geoscientific Model Development}{10}{6}{2321--2332}.
\PrintBackRefs{\CurrentBib}

\bibitem [\protect \citeauthoryear {%
Gneiting%
, Raftery%
, Westveld%
\BCBL {}\ \BBA {} Goldman%
}{%
Gneiting%
\ \protect \BOthers {.}}{%
{\protect \APACyear {2005}}%
}]{%
Gneiting2005}
\APACinsertmetastar {%
Gneiting2005}%
\begin{APACrefauthors}%
Gneiting, T.%
, Raftery, A\BPBI E.%
, Westveld, A\BPBI H.%
\BCBL {}\ \BBA {} Goldman, T.%
\end{APACrefauthors}%
\unskip\
\newblock
\APACrefYearMonthDay{2005}{}{}.
\newblock
{\BBOQ}\APACrefatitle {Calibrated Probabilistic Forecasting Using Ensemble Model Output Statistics and Minimum {{CRPS}} Estimation} {Calibrated probabilistic forecasting using ensemble model output statistics and minimum {{CRPS}} estimation}.{\BBCQ}
\newblock
\APACjournalVolNumPages{Monthly Weather Review}{133}{5}{1098--1118}.
\newblock
\begin{APACrefDOI} \doi{10.1175/MWR2904.1} \end{APACrefDOI}
\PrintBackRefs{\CurrentBib}

\bibitem [\protect \citeauthoryear {%
Hadigol%
, Maute%
\BCBL {}\ \BBA {} Doostan%
}{%
Hadigol%
\ \protect \BOthers {.}}{%
{\protect \APACyear {2015}}%
}]{%
hadigol2015}
\APACinsertmetastar {%
hadigol2015}%
\begin{APACrefauthors}%
Hadigol, M.%
, Maute, K.%
\BCBL {}\ \BBA {} Doostan, A.%
\end{APACrefauthors}%
\unskip\
\newblock
\APACrefYearMonthDay{2015}{}{}.
\newblock
{\BBOQ}\APACrefatitle {On uncertainty quantification of lithium-ion batteries: Application to an LiC6/LiCoO2 cell} {On uncertainty quantification of lithium-ion batteries: Application to an lic6/licoo2 cell}.{\BBCQ}
\newblock
\APACjournalVolNumPages{Journal of Power Sources}{300}{}{507--524}.
\PrintBackRefs{\CurrentBib}

\bibitem [\protect \citeauthoryear {%
Hodson%
}{%
Hodson%
}{%
{\protect \APACyear {2022}}%
}]{%
hodson2022}
\APACinsertmetastar {%
hodson2022}%
\begin{APACrefauthors}%
Hodson, T\BPBI O.%
\end{APACrefauthors}%
\unskip\
\newblock
\APACrefYearMonthDay{2022}{}{}.
\newblock
{\BBOQ}\APACrefatitle {Root mean square error (RMSE) or mean absolute error (MAE): When to use them or not} {Root mean square error (rmse) or mean absolute error (mae): When to use them or not}.{\BBCQ}
\newblock
\APACjournalVolNumPages{Geoscientific Model Development Discussions}{2022}{}{1--10}.
\PrintBackRefs{\CurrentBib}

\bibitem [\protect \citeauthoryear {%
Jung%
, Jo%
, Choo%
\BCBL {}\ \BBA {} Lee%
}{%
Jung%
\ \protect \BOthers {.}}{%
{\protect \APACyear {2022}}%
}]{%
jung2022}
\APACinsertmetastar {%
jung2022}%
\begin{APACrefauthors}%
Jung, Y.%
, Jo, H.%
, Choo, J.%
\BCBL {}\ \BBA {} Lee, I.%
\end{APACrefauthors}%
\unskip\
\newblock
\APACrefYearMonthDay{2022}{{\APACmonth{06}}}{}.
\newblock
{\BBOQ}\APACrefatitle {Statistical Model Calibration and Design Optimization under Aleatory and Epistemic Uncertainty} {Statistical model calibration and design optimization under aleatory and epistemic uncertainty}.{\BBCQ}
\newblock
\APACjournalVolNumPages{Reliability Engineering \& System Safety}{222}{}{108428}.
\newblock
\begin{APACrefDOI} \doi{10.1016/j.ress.2022.108428} \end{APACrefDOI}
\PrintBackRefs{\CurrentBib}

\bibitem [\protect \citeauthoryear {%
Kuleshov%
, Fenner%
\BCBL {}\ \BBA {} Ermon%
}{%
Kuleshov%
\ \protect \BOthers {.}}{%
{\protect \APACyear {2018}}%
}]{%
kuleshov2018}
\APACinsertmetastar {%
kuleshov2018}%
\begin{APACrefauthors}%
Kuleshov, V.%
, Fenner, N.%
\BCBL {}\ \BBA {} Ermon, S.%
\end{APACrefauthors}%
\unskip\
\newblock
\APACrefYearMonthDay{2018}{{\APACmonth{07}}}{}.
\newblock
{\BBOQ}\APACrefatitle {Accurate Uncertainties for Deep Learning Using Calibrated Regression} {Accurate uncertainties for deep learning using calibrated regression}.{\BBCQ}
\newblock
\BIn{} J.~Dy\ \BBA {} A.~Krause\ (\BEDS), \APACrefbtitle {Proceedings of the 35th International Conference on Machine Learning} {Proceedings of the 35th international conference on machine learning}\ (\BVOL~80, \BPGS\ 2796--2804).
\newblock
\APACaddressPublisher{}{{PMLR}}.
\PrintBackRefs{\CurrentBib}

\bibitem [\protect \citeauthoryear {%
LeBlanc%
\ \BBA {} Tibshirani%
}{%
LeBlanc%
\ \BBA {} Tibshirani%
}{%
{\protect \APACyear {1996}}%
}]{%
leblanc1996}
\APACinsertmetastar {%
leblanc1996}%
\begin{APACrefauthors}%
LeBlanc, M.%
\BCBT {}\ \BBA {} Tibshirani, R.%
\end{APACrefauthors}%
\unskip\
\newblock
\APACrefYearMonthDay{1996}{}{}.
\newblock
{\BBOQ}\APACrefatitle {Combining estimates in regression and classification} {Combining estimates in regression and classification}.{\BBCQ}
\newblock
\APACjournalVolNumPages{Journal of the American Statistical Association}{91}{436}{1641--1650}.
\PrintBackRefs{\CurrentBib}

\bibitem [\protect \citeauthoryear {%
Lee%
, Kwon%
\BCBL {}\ \BBA {} Lee%
}{%
Lee%
\ \protect \BOthers {.}}{%
{\protect \APACyear {2023}}%
}]{%
LEE2023}
\APACinsertmetastar {%
LEE2023}%
\begin{APACrefauthors}%
Lee, G.%
, Kwon, D.%
\BCBL {}\ \BBA {} Lee, C.%
\end{APACrefauthors}%
\unskip\
\newblock
\APACrefYearMonthDay{2023}{}{}.
\newblock
{\BBOQ}\APACrefatitle {A Convolutional Neural Network Model for {{SOH}} Estimation of {{Li-ion}} Batteries with Physical Interpretability} {A convolutional neural network model for {{SOH}} estimation of {{Li-ion}} batteries with physical interpretability}.{\BBCQ}
\newblock
\APACjournalVolNumPages{Mechanical Systems and Signal Processing}{188}{}{110004}.
\newblock
\begin{APACrefDOI} \doi{10.1016/j.ymssp.2022.110004} \end{APACrefDOI}
\PrintBackRefs{\CurrentBib}

\bibitem [\protect \citeauthoryear {%
Liu%
\ \protect \BOthers {.}}{%
Liu%
\ \protect \BOthers {.}}{%
{\protect \APACyear {2023}}%
}]{%
liu2023a}
\APACinsertmetastar {%
liu2023a}%
\begin{APACrefauthors}%
Liu, Y.%
, Sun, J.%
, Shang, Y.%
, Zhang, X.%
, Ren, S.%
\BCBL {}\ \BBA {} Wang, D.%
\end{APACrefauthors}%
\unskip\
\newblock
\APACrefYearMonthDay{2023}{{\APACmonth{05}}}{}.
\newblock
{\BBOQ}\APACrefatitle {A Novel Remaining Useful Life Prediction Method for Lithium-Ion Battery Based on Long Short-Term Memory Network Optimized by Improved Sparrow Search Algorithm} {A novel remaining useful life prediction method for lithium-ion battery based on long short-term memory network optimized by improved sparrow search algorithm}.{\BBCQ}
\newblock
\APACjournalVolNumPages{Journal of Energy Storage}{61}{}{106645}.
\newblock
\begin{APACrefDOI} \doi{10.1016/j.est.2023.106645} \end{APACrefDOI}
\PrintBackRefs{\CurrentBib}

\bibitem [\protect \citeauthoryear {%
Nam%
, Yoon%
, Lee%
\BCBL {}\ \BBA {} Lee%
}{%
Nam%
\ \protect \BOthers {.}}{%
{\protect \APACyear {2021}}%
}]{%
Nam2021}
\APACinsertmetastar {%
Nam2021}%
\begin{APACrefauthors}%
Nam, G.%
, Yoon, J.%
, Lee, Y.%
\BCBL {}\ \BBA {} Lee, J.%
\end{APACrefauthors}%
\unskip\
\newblock
\APACrefYearMonthDay{2021}{}{}.
\newblock
{\BBOQ}\APACrefatitle {Diversity Matters When Learning from Ensembles} {Diversity matters when learning from ensembles}.{\BBCQ}
\newblock
\BIn{} M.~Ranzato, A.~Beygelzimer, Y.~Dauphin, P.~Liang\BCBL {}\ \BBA {} J\BPBI W.~Vaughan\ (\BEDS), \APACrefbtitle {Advances in Neural Information Processing Systems} {Advances in neural information processing systems}\ (\BVOL~34, \BPGS\ 8367--8377).
\newblock
\APACaddressPublisher{}{Curran Associates, Inc.}
\PrintBackRefs{\CurrentBib}

\bibitem [\protect \citeauthoryear {%
Nemani%
\ \protect \BOthers {.}}{%
Nemani%
\ \protect \BOthers {.}}{%
{\protect \APACyear {2023}}%
}]{%
Nemani2023}
\APACinsertmetastar {%
Nemani2023}%
\begin{APACrefauthors}%
Nemani, V.%
, Biggio, L.%
, Huan, X.%
, Hu, Z.%
, Fink, O.%
, Tran, A.%
\BDBL {}Hu, C.%
\end{APACrefauthors}%
\unskip\
\newblock
\APACrefYearMonthDay{2023}{}{}.
\newblock
{\BBOQ}\APACrefatitle {Uncertainty Quantification in Machine Learning for Engineering Design and Health Prognostics: {{A}} Tutorial} {Uncertainty quantification in machine learning for engineering design and health prognostics: {{A}} tutorial}.{\BBCQ}
\newblock
\APACjournalVolNumPages{Mechanical Systems and Signal Processing}{205}{}{110796}.
\newblock
\begin{APACrefDOI} \doi{10.1016/j.ymssp.2023.110796} \end{APACrefDOI}
\PrintBackRefs{\CurrentBib}

\bibitem [\protect \citeauthoryear {%
Saha%
\ \BBA {} Goebel%
}{%
Saha%
\ \BBA {} Goebel%
}{%
{\protect \APACyear {2007}}%
}]{%
saha2007}
\APACinsertmetastar {%
saha2007}%
\begin{APACrefauthors}%
Saha, B.%
\BCBT {}\ \BBA {} Goebel, K.%
\end{APACrefauthors}%
\unskip\
\newblock
\APACrefYearMonthDay{2007}{}{}.
\newblock
{\BBOQ}\APACrefatitle {NASA Ames prognostics data repository} {Nasa ames prognostics data repository}.{\BBCQ}
\newblock
\APACjournalVolNumPages{NASA Ames, Moffett Field, CA, USA}{}{}{}.
\PrintBackRefs{\CurrentBib}

\bibitem [\protect \citeauthoryear {%
Toughzaoui%
\ \protect \BOthers {.}}{%
Toughzaoui%
\ \protect \BOthers {.}}{%
{\protect \APACyear {2022}}%
}]{%
TOUGHZAOUI2022}
\APACinsertmetastar {%
TOUGHZAOUI2022}%
\begin{APACrefauthors}%
Toughzaoui, Y.%
, Toosi, S\BPBI B.%
, Chaoui, H.%
, Louahlia, H.%
, Petrone, R.%
, Le~Masson, S.%
\BCBL {}\ \BBA {} Gualous, H.%
\end{APACrefauthors}%
\unskip\
\newblock
\APACrefYearMonthDay{2022}{}{}.
\newblock
{\BBOQ}\APACrefatitle {State of Health Estimation and Remaining Useful Life Assessment of Lithium-Ion Batteries: {{A}} Comparative Study} {State of health estimation and remaining useful life assessment of lithium-ion batteries: {{A}} comparative study}.{\BBCQ}
\newblock
\APACjournalVolNumPages{Journal of Energy Storage}{51}{}{104520}.
\newblock
\begin{APACrefDOI} \doi{10.1016/j.est.2022.104520} \end{APACrefDOI}
\PrintBackRefs{\CurrentBib}

\bibitem [\protect \citeauthoryear {%
Tran%
\ \protect \BOthers {.}}{%
Tran%
\ \protect \BOthers {.}}{%
{\protect \APACyear {2020}}%
}]{%
Tran_2020}
\APACinsertmetastar {%
Tran_2020}%
\begin{APACrefauthors}%
Tran, K.%
, Neiswanger, W.%
, Yoon, J.%
, Zhang, Q.%
, Xing, E.%
\BCBL {}\ \BBA {} Ulissi, Z\BPBI W.%
\end{APACrefauthors}%
\unskip\
\newblock
\APACrefYearMonthDay{2020}{{\APACmonth{05}}}{}.
\newblock
{\BBOQ}\APACrefatitle {Methods for Comparing Uncertainty Quantifications for Material Property Predictions} {Methods for comparing uncertainty quantifications for material property predictions}.{\BBCQ}
\newblock
\APACjournalVolNumPages{Machine Learning: Science and Technology}{1}{2}{025006}.
\newblock
\begin{APACrefDOI} \doi{10.1088/2632-2153/ab7e1a} \end{APACrefDOI}
\PrintBackRefs{\CurrentBib}

\bibitem [\protect \citeauthoryear {%
Vanem%
, Salucci%
, Bakdi%
\BCBL {}\ \BBA {} Alnes%
}{%
Vanem%
\ \protect \BOthers {.}}{%
{\protect \APACyear {2021}}%
}]{%
vanem2021}
\APACinsertmetastar {%
vanem2021}%
\begin{APACrefauthors}%
Vanem, E.%
, Salucci, C\BPBI B.%
, Bakdi, A.%
\BCBL {}\ \BBA {} Alnes, {\O}\BPBI {\AA}\BPBI S.%
\end{APACrefauthors}%
\unskip\
\newblock
\APACrefYearMonthDay{2021}{{\APACmonth{11}}}{}.
\newblock
{\BBOQ}\APACrefatitle {Data-Driven State of Health Modelling---{{A}} Review of State of the Art and Reflections on Applications for Maritime Battery Systems} {Data-driven state of health modelling---{{A}} review of state of the art and reflections on applications for maritime battery systems}.{\BBCQ}
\newblock
\APACjournalVolNumPages{Journal of Energy Storage}{43}{}{103158}.
\newblock
\begin{APACrefDOI} \doi{10.1016/j.est.2021.103158} \end{APACrefDOI}
\PrintBackRefs{\CurrentBib}

\bibitem [\protect \citeauthoryear {%
Wang%
\ \protect \BOthers {.}}{%
Wang%
\ \protect \BOthers {.}}{%
{\protect \APACyear {2022}}%
}]{%
wang2022}
\APACinsertmetastar {%
wang2022}%
\begin{APACrefauthors}%
Wang, C\BHBI j.%
, Zhu, Y\BHBI l.%
, Gao, F.%
, Bu, X\BHBI y.%
, Chen, H\BHBI s.%
, Quan, T.%
\BDBL {}Jiao, Q\BHBI j.%
\end{APACrefauthors}%
\unskip\
\newblock
\APACrefYearMonthDay{2022}{}{}.
\newblock
{\BBOQ}\APACrefatitle {Internal short circuit and thermal runaway evolution mechanism of fresh and retired lithium-ion batteries with LiFePO4 cathode during overcharge} {Internal short circuit and thermal runaway evolution mechanism of fresh and retired lithium-ion batteries with lifepo4 cathode during overcharge}.{\BBCQ}
\newblock
\APACjournalVolNumPages{Applied Energy}{328}{}{120224}.
\PrintBackRefs{\CurrentBib}

\bibitem [\protect \citeauthoryear {%
Wei%
\ \BBA {} Wu%
}{%
Wei%
\ \BBA {} Wu%
}{%
{\protect \APACyear {2023}}%
}]{%
wei2023}
\APACinsertmetastar {%
wei2023}%
\begin{APACrefauthors}%
Wei, Y.%
\BCBT {}\ \BBA {} Wu, D.%
\end{APACrefauthors}%
\unskip\
\newblock
\APACrefYearMonthDay{2023}{{\APACmonth{02}}}{}.
\newblock
{\BBOQ}\APACrefatitle {Prediction of State of Health and Remaining Useful Life of Lithium-Ion Battery Using Graph Convolutional Network with Dual Attention Mechanisms} {Prediction of state of health and remaining useful life of lithium-ion battery using graph convolutional network with dual attention mechanisms}.{\BBCQ}
\newblock
\APACjournalVolNumPages{Reliability Engineering \& System Safety}{230}{}{108947}.
\newblock
\begin{APACrefDOI} \doi{10.1016/j.ress.2022.108947} \end{APACrefDOI}
\PrintBackRefs{\CurrentBib}

\bibitem [\protect \citeauthoryear {%
Yang%
, Chen%
, Chen%
\BCBL {}\ \BBA {} Huang%
}{%
Yang%
\ \protect \BOthers {.}}{%
{\protect \APACyear {2023}}%
}]{%
yang2023}
\APACinsertmetastar {%
yang2023}%
\begin{APACrefauthors}%
Yang, Y.%
, Chen, S.%
, Chen, T.%
\BCBL {}\ \BBA {} Huang, L.%
\end{APACrefauthors}%
\unskip\
\newblock
\APACrefYearMonthDay{2023}{}{}.
\newblock
{\BBOQ}\APACrefatitle {State of Health Assessment of Lithium-Ion Batteries Based on Deep Gaussian Process Regression Considering Heterogeneous Features} {State of health assessment of lithium-ion batteries based on deep gaussian process regression considering heterogeneous features}.{\BBCQ}
\newblock
\APACjournalVolNumPages{Journal of Energy Storage}{61}{}{106797}.
\PrintBackRefs{\CurrentBib}

\bibitem [\protect \citeauthoryear {%
Yao%
, Vehtari%
, Simpson%
\BCBL {}\ \BBA {} Gelman%
}{%
Yao%
\ \protect \BOthers {.}}{%
{\protect \APACyear {2018}}%
}]{%
yao2018}
\APACinsertmetastar {%
yao2018}%
\begin{APACrefauthors}%
Yao, Y.%
, Vehtari, A.%
, Simpson, D.%
\BCBL {}\ \BBA {} Gelman, A.%
\end{APACrefauthors}%
\unskip\
\newblock
\APACrefYearMonthDay{2018}{{\APACmonth{09}}}{}.
\newblock
{\BBOQ}\APACrefatitle {Using {{Stacking}} to {{Average Bayesian Predictive Distributions}} (with {{Discussion}})} {Using {{Stacking}} to {{Average Bayesian Predictive Distributions}} (with {{Discussion}})}.{\BBCQ}
\newblock
\APACjournalVolNumPages{Bayesian Analysis}{13}{3}{}.
\newblock
\begin{APACrefDOI} \doi{10.1214/17-BA1091} \end{APACrefDOI}
\PrintBackRefs{\CurrentBib}

\bibitem [\protect \citeauthoryear {%
Zamo%
\ \BBA {} Naveau%
}{%
Zamo%
\ \BBA {} Naveau%
}{%
{\protect \APACyear {2018}}%
}]{%
zamo2018}
\APACinsertmetastar {%
zamo2018}%
\begin{APACrefauthors}%
Zamo, M.%
\BCBT {}\ \BBA {} Naveau, P.%
\end{APACrefauthors}%
\unskip\
\newblock
\APACrefYearMonthDay{2018}{{\APACmonth{02}}}{}.
\newblock
{\BBOQ}\APACrefatitle {Estimation of the {{Continuous Ranked Probability Score}} with {{Limited Information}} and {{Applications}} to {{Ensemble Weather Forecasts}}} {Estimation of the {{Continuous Ranked Probability Score}} with {{Limited Information}} and {{Applications}} to {{Ensemble Weather Forecasts}}}.{\BBCQ}
\newblock
\APACjournalVolNumPages{Mathematical Geosciences}{50}{2}{209--234}.
\newblock
\begin{APACrefDOI} \doi{10.1007/s11004-017-9709-7} \end{APACrefDOI}
\PrintBackRefs{\CurrentBib}

\bibitem [\protect \citeauthoryear {%
S.~Zhang%
, Liu%
\BCBL {}\ \BBA {} Su%
}{%
S.~Zhang%
\ \protect \BOthers {.}}{%
{\protect \APACyear {2022}}%
}]{%
Zhang2022}
\APACinsertmetastar {%
Zhang2022}%
\begin{APACrefauthors}%
Zhang, S.%
, Liu, Z.%
\BCBL {}\ \BBA {} Su, H.%
\end{APACrefauthors}%
\unskip\
\newblock
\APACrefYearMonthDay{2022}{}{}.
\newblock
{\BBOQ}\APACrefatitle {A Bayesian Mixture Neural Network for Remaining Useful Life Prediction of Lithium-Ion Batteries} {A bayesian mixture neural network for remaining useful life prediction of lithium-ion batteries}.{\BBCQ}
\newblock
\APACjournalVolNumPages{IEEE Transactions on Transportation Electrification}{8}{4}{4708--4721}.
\newblock
\begin{APACrefDOI} \doi{10.1109/TTE.2022.3161140} \end{APACrefDOI}
\PrintBackRefs{\CurrentBib}

\bibitem [\protect \citeauthoryear {%
Y.~Zhang%
, Zhang%
, Liu%
, Feng%
\BCBL {}\ \BBA {} Xu%
}{%
Y.~Zhang%
\ \protect \BOthers {.}}{%
{\protect \APACyear {2024}}%
}]{%
zhang2024}
\APACinsertmetastar {%
zhang2024}%
\begin{APACrefauthors}%
Zhang, Y.%
, Zhang, M.%
, Liu, C.%
, Feng, Z.%
\BCBL {}\ \BBA {} Xu, Y.%
\end{APACrefauthors}%
\unskip\
\newblock
\APACrefYearMonthDay{2024}{{\APACmonth{02}}}{}.
\newblock
{\BBOQ}\APACrefatitle {Reliability Enhancement of State of Health Assessment Model of Lithium-Ion Battery Considering the Uncertainty with Quantile Distribution of Deep Features} {Reliability enhancement of state of health assessment model of lithium-ion battery considering the uncertainty with quantile distribution of deep features}.{\BBCQ}
\newblock
\APACjournalVolNumPages{Reliability Engineering \& System Safety}{}{}{110002}.
\newblock
\begin{APACrefDOI} \doi{10.1016/j.ress.2024.110002} \end{APACrefDOI}
\PrintBackRefs{\CurrentBib}

\bibitem [\protect \citeauthoryear {%
H.~Zhao%
\ \protect \BOthers {.}}{%
H.~Zhao%
\ \protect \BOthers {.}}{%
{\protect \APACyear {2023}}%
}]{%
ZHAO2023}
\APACinsertmetastar {%
ZHAO2023}%
\begin{APACrefauthors}%
Zhao, H.%
, Chen, Z.%
, Shu, X.%
, Shen, J.%
, Lei, Z.%
\BCBL {}\ \BBA {} Zhang, Y.%
\end{APACrefauthors}%
\unskip\
\newblock
\APACrefYearMonthDay{2023}{}{}.
\newblock
{\BBOQ}\APACrefatitle {State of Health Estimation for Lithium-Ion Batteries Based on Hybrid Attention and Deep Learning} {State of health estimation for lithium-ion batteries based on hybrid attention and deep learning}.{\BBCQ}
\newblock
\APACjournalVolNumPages{Reliability Engineering \& System Safety}{232}{}{109066}.
\newblock
\begin{APACrefDOI} \doi{10.1016/j.ress.2022.109066} \end{APACrefDOI}
\PrintBackRefs{\CurrentBib}

\bibitem [\protect \citeauthoryear {%
X.~Zhao%
, Wang%
, Li%
\BCBL {}\ \BBA {} Miao%
}{%
X.~Zhao%
\ \protect \BOthers {.}}{%
{\protect \APACyear {2024}}%
}]{%
zhao2024}
\APACinsertmetastar {%
zhao2024}%
\begin{APACrefauthors}%
Zhao, X.%
, Wang, Z.%
, Li, E.%
\BCBL {}\ \BBA {} Miao, H.%
\end{APACrefauthors}%
\unskip\
\newblock
\APACrefYearMonthDay{2024}{{\APACmonth{01}}}{}.
\newblock
{\BBOQ}\APACrefatitle {Investigation into {{Impedance Measurements}} for {{Rapid Capacity Estimation}} of {{Lithium-ion Batteries}} in {{Electric Vehicles}}} {Investigation into {{Impedance Measurements}} for {{Rapid Capacity Estimation}} of {{Lithium-ion Batteries}} in {{Electric Vehicles}}}.{\BBCQ}
\newblock
\APACjournalVolNumPages{Journal of Dynamics, Monitoring and Diagnostics}{}{}{}.
\newblock
\begin{APACrefDOI} \doi{10.37965/jdmd.2024.475} \end{APACrefDOI}
\PrintBackRefs{\CurrentBib}

\end{thebibliography}

\end{document}